\documentclass[11pt]{article}
\usepackage{acl}

\usepackage{times}
\usepackage{latexsym}

\usepackage[utf8]{inputenc}

\usepackage{microtype}
\usepackage{url}
\usepackage{amsmath}
\usepackage{amsthm}
\usepackage[inline]{enumitem}
\usepackage{multirow}
\usepackage{subfig}
\usepackage{color, colortbl}
\usepackage{inconsolata}
\usepackage{xspace}
\usepackage{floatrow}
\usepackage{algorithm}
\usepackage{algpseudocode}
\usepackage{soul}
\usepackage{fancyvrb}
\usepackage{listings}
\usepackage{booktabs}
\usepackage{graphicx}
\usepackage{amsfonts}

\newfloatcommand{capbtabbox}{table}[][\FBwidth]
\newfloat{lstfloat}{htbp}{lop}
\floatname{lstfloat}{Prompt}

\newfloat{lstdoc}{htbp}{lop}
\floatname{lstdoc}{Guideline}

\makeatletter
\def\@fnsymbol#1{\ensuremath{\ifcase#1\or \dagger\or \ddagger\or
   \mathsection\or \mathparagraph\or \|\or **\or \dagger\dagger
   \or \ddagger\ddagger \else\@ctrerr\fi}}

\title{\textsc{Pearl}: Personalizing Large Language Model Writing \\ Assistants with Generation-Calibrated Retrievers}

\author{Sheshera Mysore\textsuperscript{1}$^{\bigtriangleup\dagger}$\quad
Zhuoran Lu\textsuperscript{2}\thanks{~~Work done during internship at Microsoft Research.}\quad
Mengting Wan\textsuperscript{1}\quad 
{\bf Longqi Yang\textsuperscript{1}}\quad\\
{\bf Bahar Sarrafzadeh\textsuperscript{1}}\quad
{\bf Steve Menezes\textsuperscript{1}}\quad
{\bf Tina Baghaee\textsuperscript{1}}\quad\\
{\bf Emmanuel Barajas Gonzalez\textsuperscript{1}}\quad
{\bf Jennifer Neville\textsuperscript{1}}\quad
{\bf Tara Safavi\textsuperscript{1}$^\bigtriangleup$}\\
\textsuperscript{2}Purdue University, \textsc{\small IN, USA}
\textsuperscript{1}Microsoft, \textsc{\small WA, USA}\\
  $^\bigtriangleup$Corresponding authors: \texttt{smysore@cs.umass.edu}, \texttt{tarasafavi@microsoft.com} \\
  }

\begin{document}
\newcommand{\tara}[1]{\textcolor{red}{[Tara: #1]}}
\newcommand{\proposedmethod}{\textsc{Pearl}\xspace} 
\newcommand{\worksm}{\textsc{WorkSm}\xspace}
\newcommand{\aita}{\textsc{AITA}\xspace}
\newcommand*\cirnum[1]{\raisebox{.5pt}{\textcircled{\raisebox{-.9pt} {#1}}}}
\newcommand{\historicalposts}{\mathcal{D}_u}
\newcommand{\llmG}{f_\textrm{LLM}}
\newcommand{\slm}{f_\textrm{aux}}
\newcommand{\retriever}{f_{\textrm{retr}}}
\newcommand{\request}{q_u}
\newcommand{\doc}{d_u}
\newcommand{\target}{t_u}
\maketitle
\begin{abstract}
Powerful large language models have facilitated the development of writing assistants that promise to significantly improve the quality and efficiency of composition and communication.
However, a barrier to effective assistance is the lack of personalization in LLM outputs to the author's communication style, specialized knowledge, and values. 
In this paper, we address this challenge by proposing \proposedmethod, a LLM writing assistant personalized with a retriever that is trained to be \emph{generation-calibrated} for personalization. Generation calibration ensures that our retriever selects historic user authored documents to augment an LLM prompt such that they are likely to help an LLM generation better adhere to a users' preferences. 
We propose two key novelties for training such a retriever: (1) A training data selection method that identifies historical user requests likely to benefit from personalization \textit{and} documents that provide that benefit; and (2) A scale-calibrating KL-divergence objective that ensures that our retriever scores remain proportional to the downstream generation quality from using the document for personalized generation. In a series of holistic evaluations, we demonstrate the effectiveness of \proposedmethod in generating long-form texts on multiple social media datasets. Finally, we demonstrate how a generation-calibrated retriever can double as a performance predictor -- detecting low quality retrieval, and improving potentially under-performing outputs via revision with LLMs.
\end{abstract}


\section{Introduction}
\label{sec:intro}
Machine-assisted writing has seen a long history of development, progressing from providing simple syntactic checks, to revising human authored text, to recent assistants being able to fully compose texts on direction from authors \citep{mahlow2023writing, dale_viethen_2021}. The text-generation capabilities of current LLMs and has led current research to explore a new frontier of writing assistants for complex applications such as knowledge synthesis \citep{shen2023summarization}, peer review \citep{chen2023peerwrite}, and journalism \citep{wang2023writingjourn}. 
An important element of effective writing assistants is being able to {personalize} generated text to retain the knowledge, style, and values of a user -- an essential element of interpersonal communication \citep{pickering2013integrated}.  
With current LLMs prone to generating overly generic text \citep{pu2023gptvhuman}, author personalization of LLMs is an important problem.

\begin{figure}[t]
     \centering
     {\includegraphics[width=0.7\textwidth]{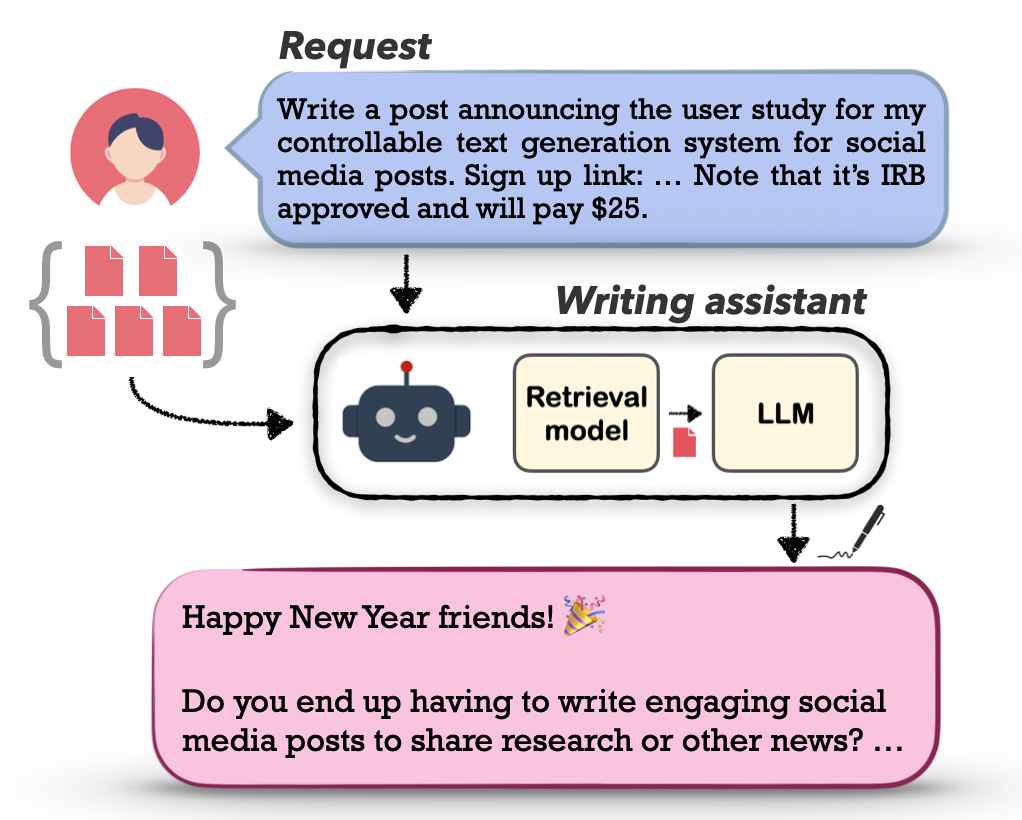}}
     \caption{\proposedmethod is a request-driven generation model that personalizes LLM outputs through retrieval augmentation with a \emph{generation calibrated} retriever.}
     \label{fig-teaser}
 \end{figure}

Personalizing LLM outputs may be seen as a form of alignment to individual users of the LLM \citep{kirk2023personalisation}. However, leveraging fine-tuning for alignment in a personalization setup poses challenges to serving trained per-user models and obtaining sufficient per-user alignment training data. Therefore, we pursue in-context alignment through retrieval augmentation  \citep{salemi2023lamp, li2023automatic}. First, we assume access to a set of historic \emph{user-authored documents} (e.g.\ emails, social media posts, etc.) and a user \emph{request} for a personalized generation. To personalize LLM outputs we propose an approach to train a retrieval model that selects historic user documents to augment an LLM's prompt. Historic documents capture users' personal style, knowledge, and values and can serve as useful context for personalized generation. 
While training retrievers for non-personalized applications have been explored in prior work \cite{gonen2022demystifying}, this exploration has been limited in personalized text generation. Finally, we pursue personalization of LLMs only accessible via prompt-based APIs since this represents a common form of accessing performant large scale LLMs.

The starting point for our retriever in prior work examining effective prompts for \emph{non-personalized} applications:  \citet{gonen2022demystifying} show the best prompts to be those with the highest conditional likelihood of generating a target text, and \citet{rubin2022promptret} use these likelihoods to train retrieval models for non-personalized retrieval augmentation of LLMs.
While this approach performs well in non-personalized setups, \emph{personalized} text generation presents unique challenges and opportunities: There are fewer historic documents per user ($\sim$hundreds) than common non-personalized retrieval collections, and user requests may diverge from their history as users' preferences change. A smaller retrieval corpus and shifting interests mean that all requests cannot be satisfied by retrieval from a users' historical documents -- as a result, all historic requests and documents are unlikely to be useful for training a retriever.
Our first contribution addresses this: We present a novel \textbf{difference of likelihoods}-based method that identifies \emph{only} the personalizable user requests and associated documents that are likely to personalize downstream generations, and use these to train our retriever.

Next, the personalization setup offers an opportunity: Fewer historical documents per user permits the use of expressive cross-encoder retrievers instead of scalable but less expressive biencoders commonly used for non-personalized tasks \citep{rubin2022promptret}. However, cross-encoders produce skewed scores at the ends of their score ranges \citep{menon22defensedual, yadav2022anncur}, hampering their ability to closely track the utility of a document for personalized generation. We remedy this with our second contribution -- a \textbf{personalized scale-calibrating training objective} \citep{le2022scalecalib}. This ensures that scores from our retriever are \emph{generation-calibrated} for personalization -- i.e.\ the score it produces for request-document pairs is proportional to the output quality of an LLM prompted with the pair. In a case study, we show how generation calibration enables the retriever's scores to be used for \emph{retrieval performance prediction} -- detecting low-quality retrievals, and revising potentially low-quality generations.

We instantiate \proposedmethod with multiple LLMs, \texttt{davinci-003} and \texttt{gpt-35-turbo}, at privacy compliant enterprise API endpoints and evaluate it on a private dataset of workplace communications and a public dataset of Reddit comments.
For evaluation, we use a variety of evaluation methods spanning intrinsic, extrinsic, and personalized LLM-as-judge evaluations to demonstrate the value of \proposedmethod. Further, since we train calibrated retrieval models, we present additional evaluations for calibration, ablations, and analysis in Appendices. 
Our evaluations demonstrate that \proposedmethod consistently matches or outperforms strong baseline approaches. 
\begin{figure*}[t]
     \centering
     \fbox{\includegraphics[width=0.7\textwidth]{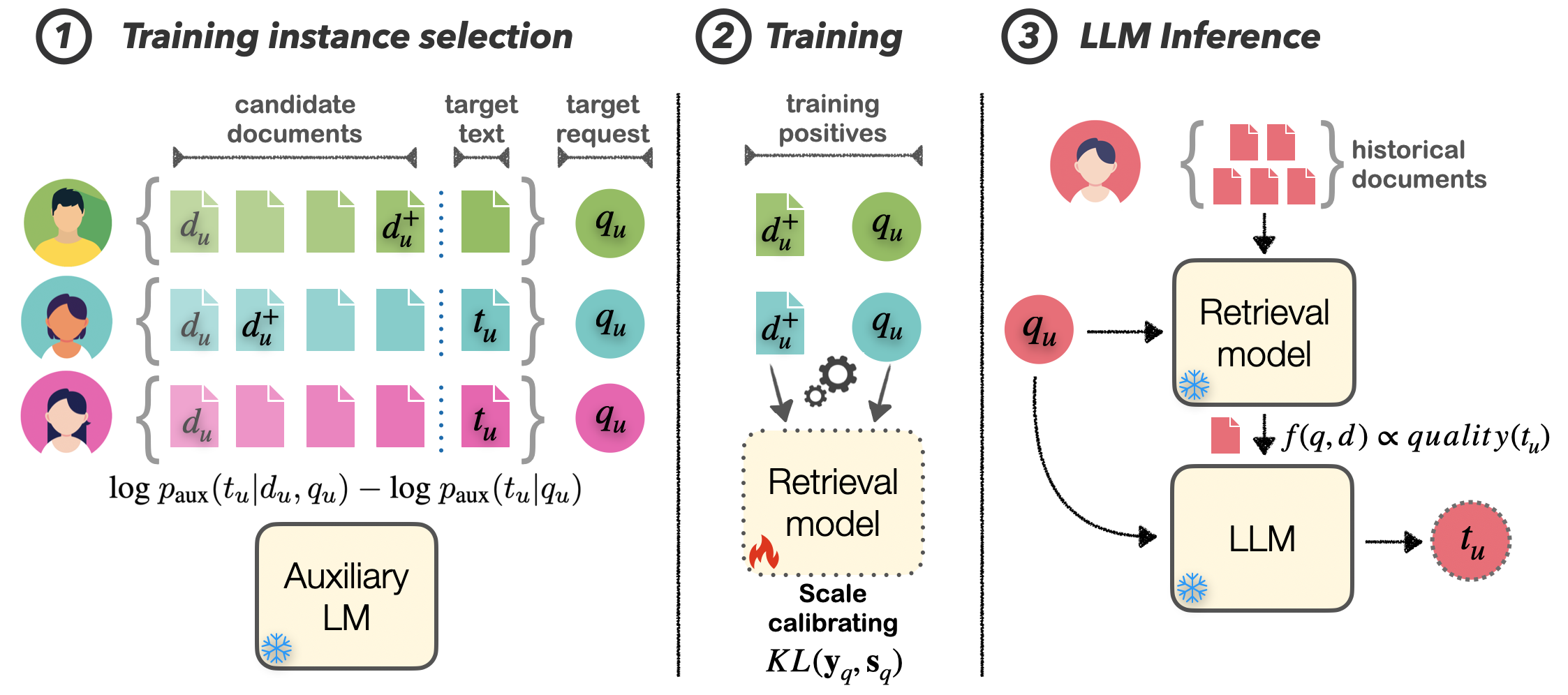}}
     \caption{To train retriever, $\retriever$, an auxiliary language model is first used to identify historical \textit{requests} that can be personalized and the best \textit{document} to use for personalization \cirnum{\footnotesize 1}. Then, $\retriever$ is trained on the selected data with a scale calibrating loss function \cirnum{\footnotesize 2}. Given an unseen request, $\retriever$ is used to select the best instances from historical texts for augmenting an LLM prompt for personalized generation \cirnum{\footnotesize 3}. Our training results in a generation calibrated retriever where scores for documents are proportional to the quality of the LLM output.}
     \label{fig-high-level}
 \end{figure*}

\section{Related Work}
\label{sec:related}

\textbf{Example selection for LLMs} 
Early work on training retrievers for augmenting LLM contexts in non-personalized applications was proposed by \citet{rubin2022promptret}. They train retrieval models by distilling LLM likelihoods of the target completions conditioned on the prompt. Similarly \citet{wang2023learning} train retrieval models on finer-grained feedback from a trained reward model through  distillation. More distantly, \citet{zhang2022active} train instances selection models on rewards from a downstream evaluation metric using reinforcement learning. Parallel with our work, \citet{salemi2024ropg} train bi-encoders for personalized classification and short text generation and find knowledge distillation from downstream LLMs to outperform reinforcement learning based training of retrievers. In this regard, \citet{salemi2024ropg} and \citet{rubin2022promptret} are closely related and represent closest work to ours -- we compare to such an approach in ablations (Appendix \ref{sec-ablations}). Despite similarities to our work, all prior work has explored training retrievers for document selection while assuming that satisfactory predictions can be made for \emph{all} inputs/requests. In addition to selecting documents for training, we also select training requests that benefit from retrieval augmentation -- a necessity in personalization where retrieval is performed over a smaller historical document set instead of a large shared corpus. Further, no prior approaches explore calibration for retrievers and their ability to identify low-quality retrievals, and selectively revise LLM outputs -- we explore this. Appendix \ref{sec-related-work-extended} discusses additional work on optimizing prompts, robustness to prompt errors, and calibrated retrievers.

\textbf{Personalized writing assistants} 
While writing assistants have seen considerable exploration, only some prior work has focused on author personalization. These applications range from email \citep{chen2019gmail, trajanovski2021microsoftcontext}, to social media \citep{gero2022sparks}, and grammatical error correction (GEC) \citep{nadejde2019personalgec}. These systems commonly leverage nearest-neighbor models \citep{chen2019gmail, trajanovski2021microsoftcontext} and user-group level parameter-efficient fine-tuning for personalization \citep{nadejde2019personalgec}. In contrast, we explore retrieval models for in-context alignment/personalization with LLMs. 
Parallel work has also explored personalized writing with LLMs. \citet{li2023teach} construct prompts with pre-trained retrieval and summarization models and fine-tune an LLM for personalized completion. Follow-on work has explored training a prompt-re-writer to tune prompts for a fixed LLM \citep{li2023automatic}. Prompt re-writing is a complementary approach to a trained retriever, with future systems likely to benefit from both. Appendix \ref{sec-related-work-extended} discusses non-personalized writing assistants and reader personalization.

\section{Problem Definition}
\label{sec:problem}
We consider a request-conditional, personalized text generation task. 
As input to the system, we assume a \textbf{user} $u$ who is associated with a set of $N_u$ \textbf{historical documents} $\historicalposts = \{\doc^{(i)}\}_{i=1}^{N_u}$, where each document $\doc$ may be a previously-authored social media post, email, etc. 
The user $u$ is further associated with a textual \textbf{request} $\request$ submitted to the writing assistant. The request may be authored by the user or constructed from the task context. Explicitly authored requests are increasingly common in conversational LLM interfaces \citep{papenmeier2021convsearch}, and task contexts may be seen as implicit requests e.g.\ email prefixes that require completion \citep{chen2019gmail}. 
Finally, we assume access to a \textbf{large language model}  $\llmG$ available via a prompt-based text generation API.

Given $\historicalposts$, $\request$, and $\llmG$, our retriever, $\retriever$ is trained to select a subset of historical documents $\historicalposts' \subset \historicalposts$ as few-shot examples for the LLM. Then the LLM generates a \textbf{target text} $\target$ of up to 300 words: $\target = \llmG(\phi(\request, \historicalposts'))$, where $\phi$ is a prompt construction function that inputs the user's request and retrieved historical documents, $\target$ reflects the style, knowledge, and values of $u$.

\section{Proposed Approach}
\label{sec:method}
We present \proposedmethod, an in-context aligned LLM-based model for personalized writing assistance. 
Our approach (Figure \ref{fig-high-level}) consists of an offline retriever training stage and an online LLM inference stage.
Offline, we train a \textbf{retriever} $\retriever : (\request, \doc) \rightarrow \mathbb{R}$  that scores the user's historical documents for their ability to personalize the output for a user request. Further, we ensure that $\retriever$ is generation calibrated i.e.\ the scores it produces for $(\request, \doc)$ pairs are proportional to the quality of the generated text from using $(\request, \doc)$ in a prompt. 
We train such a retriever through two key novelties: (1) Training data selection based on a novel difference of likelihoods from an auxiliary text generation model -- we identify requests which benefit from personalization \emph{and} documents which likely help personalize a target, and (2) A scale-calibrating training objective which ensures that retrievers closely track the benefit of request-document pairs for generation.
Given a new request, our LLM is prompted to generate a target text $\target$ conditioned on the request and the documents retrieved by $\retriever$. Next, we describe the retriever training set construction (Algorithm \ref{alg-train-data-selection}), how we optimize the retriever, and the details of our implementation. 
 
\subsection{Training Data Setup}
\label{method:training-data}
To optimize $\retriever$ for a personalized text generation task, we carefully create a training set for $\retriever$ from historical user documents by using an \textbf{auxiliary text generation model} $\slm$ to identify which requests and documents will help to personalize the generation of a target text.

\textbf{Data organization}
We organize the training data to create a setup close to the problem defined in \S\ref{sec:problem}.
Given a set of $M$ users and their historical document sets $\{\historicalposts\}_{u=1}^{M}$,
for each user $u$ we partition $\historicalposts$ into two non-overlapping sets, a candidate document set $\historicalposts^c \subset \historicalposts$, and a ``target'' text set $\historicalposts^t \subset \historicalposts$, such that $\historicalposts^c + \historicalposts^t = \historicalposts$.
The partitioning is done temporally, i.e.\ the target texts occur after the candidate documents, mimicking the personalization scenario where past texts are used to personalize later targets. If time data isn't available, the partitioning may be done randomly.

Next, for each target text $\target$ in each users $\historicalposts^t$, we pair the text with a corresponding request $\request$. 
For training, requests may be naturally present in the data, e.g., email prefixes that require completion \citep{chen2019gmail}, or they may be generated synthetically \citep{bonifacio2022inpars}. We detail request generation in \S\ref{exp:data}.

\textbf{Auxiliary model scoring}
Next, we use the auxiliary text generation model $\slm$ to score each candidate document in $\doc \in \historicalposts^c$ for producing the personalized $\target$ corresponding to the $\request$ for each $(\request, \target) \in \historicalposts^t$. 
We define the score as a difference in the likelihood, per $\slm$, of the target given the request with and without the historical document:
\begin{equation}
    y_{q_u}^{\doc}  = \text{log}~p_{\textrm{aux}} (\target | \doc, \request) - \text{log}~p_{\textrm{aux}}(\target | \request),
    \label{eq-diff-likelihood}
\end{equation}
Importantly, Eq.~\eqref{eq-diff-likelihood} is highest when the request is suitable for personalization \emph{and} the candidate document is the ``right'' example for personalization. That is, the request alone is not sufficient for generating the target text (i.e., the quantity defined by the second term is lower), and this candidate document is particularly beneficial to generation (i.e., the quantity defined by the first term is higher). Finally, we assume model $\slm$ to be smaller than $\llmG$ to support efficient creation of training data, and that we have access to its token likelihoods. Appendix \ref{supp-implementation} shows prompts used for $\slm$.

\subsection{Training Data Selection}
\label{method:filtering}

We use the scores from Eq.~\ref{eq-diff-likelihood} to identify: (1) a subset of training requests that are likely to benefit from personalization; and (2) candidate documents that are likely to benefit those requests i.e.\ positive training documents.

\textbf{Request selection} 
Using Eq.~\ref{eq-diff-likelihood}, we score all request-target pairs of a user in $\historicalposts^t$ against all of their candidate documents $\doc \in \historicalposts^c$, across all $M$ users. 
After scoring, we retain the top scoring $T$ request-target pairs. In practice, we find that setting $T$ to the top two-thirds across the dataset works well.
This step reflects the intuition that not all request-target pairs will benefit from retrieval augmentation, either due to the lack of suitable candidate documents in a user's historical document set, or due to underspecified requests making the target text simply too difficult to generate well -- this is contrast with RAG setups in non-personalized scenarios where a large retrieval corpus ensures that most requests are likely to benefit from retrieval.
After obtaining a high-quality set of training requests $\{q^*_u\}_{t=1}^T$, we discard the target texts, since they aren't used for training $\retriever$ or for inference.

\textbf{Candidate document selection}
Next, we use Eq.~\ref{eq-diff-likelihood} to select the best documents for the retained requests, i.e.\ identify positive training documents. 
Given a request $\request^*$ selected for training, we take the $P$ highest-scoring candidate documents $\doc \in \historicalposts^c$ as per Eq.~\eqref{eq-diff-likelihood} as positives, $\{d^+_u\}_{p=1}^P$.
We sample $N$ negative samples per positive randomly from the candidate document set for the user.
\begin{algorithm}[t]
\small
    \caption{Selecting requests and positive candidate documents to train $\retriever$}
    \begin{algorithmic}[1]
        \State \textbf{Input}: $\{\historicalposts\}_{u=1}^{M}$, $\slm$ 
        \Comment{Historical documents for $M$ users and an auxiliary LM}
        \For {each user $u$}
            \State $\historicalposts^c, \historicalposts^t \gets \mathrm{TemporalPartition}(\historicalposts)$ \Comment{Temporally partition $\historicalposts$ into candidate and target documents}
            \For {each target text $t_u \in \historicalposts^t$}
                \State $q_u \gets \mathrm{GetRequest}(t_u)$ \Comment{Obtain a synthetic or natural request}
            \EndFor
            \For {each $(\request, \target)$ pair in $\historicalposts^t$}
            \Comment{Compute benefit of personalization for request-target pairs}
                \For {each candidate $\doc$ in $\historicalposts^c$}
                    \State $Y[q_u,{\doc}]  = \text{log}~p_{\textrm{aux}} (\target | \doc, \request) - \text{log}~p_{\textrm{aux}}(\target | \request)$
                    \Comment{Equation \eqref{eq-diff-likelihood}}
                \EndFor
            \EndFor
        \EndFor
        \State $\{q^*_u\}_{t=1}^{T} \gets \mathrm{TopK}(Y[q_u,{\doc}])$ 
        \Comment{{Retain the top $T$ unique requests which are personalizable}}
        \For {each retained request in $\{q^*_u\}_{k=1}^{T}$}
            \State $\{d^+_u\}_{p=1}^{P} \gets \mathrm{TopK}(Y[q^*_u,{\doc}])$
            \Comment{{Retain the top $P$ candidates that best personalize the target}}
        \EndFor
            \State \textbf{return} $\{q^*_u, \{d^+_u\}_{p=1}^{P}\}_{t=1}^{T}$
    \end{algorithmic}
    \label{alg-train-data-selection}
\end{algorithm}

\subsection{Retriever Optimization}
\label{method:optimization}

Our $\retriever$ is a cross-encoder initialized with a pretrained LM encoder and trained using data selected per Algorithm \ref{alg-train-data-selection}, through distillation of scores in Equation \ref{eq-diff-likelihood}. While cross-encoders are expressive they produce scores which lie at the extremes of their score ranges \citep{menon22defensedual, yadav2022anncur} -- this hampers their ability to closely track the benefit of candidate documents for personalizing requests. We propose to remedy this through a scale calibrating training objective.

\textbf{Scale calibration}
Let $\mathbf{y}_{q} = [y_{q}^{+},\dots,y_{q}^{-}]$, where $y_{q}^+$ corresponds to the score of a positive document 
and $y_{q}^{-}$ corresponds to the score of a negative document from Eq.~\ref{eq-diff-likelihood}. Here, $\mathbf{y}_{q}$ contains $N$ negatives and 1 positive document.
Similarly, let the predicted logits from $\retriever : (\request, \doc) \rightarrow \mathbb{R}$ be denoted as $\mathbf{s}_{q} = [s_{q}^{+}, \dots, s_{q}^{-}]$.
Then, a standard  KL-divergence loss is written as
$KL(\mathbf{y}_{q}, \mathbf{s}_{q}) = -\sum_{i} \texttt{sm} (y_{q,i}) \text{log}~\texttt{sm} (s_{q,i})$, where $\texttt{sm}$ represents the softmax function. 
Our proposed scale calibration modifies the KL divergence loss by adding an ``anchor'' example with target score $y_0$, which is a tunable hyperparameter, and logit $s_0$ set to $0$, resulting in score vectors $\mathbf{y}_q' = [ y_0, \mathbf{y}_q ]$ and $\mathbf{s}_q'= [s_0, \mathbf{s}_q]$.
The scale-calibrated KL-divergence loss is thus
\begin{align}
    & KL(\mathbf{y}_q', \mathbf{s}_q')=-\sum_{i} \texttt{sm}(y'_{q,i})\text{log}~\texttt{sm}(s'_{q,i})
    \label{eq-kl-loss}\\
    \small
    \begin{split}
        &= - \sum_{i} \frac{e^{y_{q,i}}}{\sum_{j}e^{y_{q,j}} +e^{y_0}}\text{log}\frac{e^{s_{q,i}}}{\sum_{j}e^{s_{q,j}}+1} \\
        &+ \frac{e^{y_0}}{ \sum_{j}e^{y_{q,j}} +e^{y_0}}\text{log}~({\sum_{j}e^{s_{q,j}}+1}).
    \end{split}
    \label{eq-calibrated-expansion}
\end{align}
We find that setting $y_0$ to the median value of scores from Eq~~\eqref{eq-diff-likelihood} for positive candidate documents works well. This ensures that very large scores from $\retriever$ are penalized (second term Eq \ref{eq-calibrated-expansion}) and smaller scores are prevented from being driven lower (first term Eq \ref{eq-calibrated-expansion}). Therefore $\retriever$ scores are more evenly distributed over the score range. In practice, this ensures that predicted scores from $\retriever$ more accurately reflect the distribution of $\slm$, which in turn more closely tracks the utility of request-document pairs for personalization. We compare \proposedmethod to baselines in \S \ref{sec-main-results} and present ablations in \S \ref{sec-ablations}.

\subsection{System Details}
After training retriever $\retriever$ offline, \proposedmethod may be used to serve requests online. Given a unseen request, $\retriever$ retrieves the top-$k$ historical texts from $\historicalposts$, these are formatted into a prompt and input to $\llmG$ to generate a personalized target text $\target$.

Our $\retriever$ is initialized with a 110M parameter \textsc{MPNet} encoder~\citep{song2020mpnet}.  For $\llmG$ we consider two performant LLMs, \texttt{davinci-003} and \texttt{gpt-3.5-turbo}.
For $\slm$ we use \textsc{FlanT5-XL} with 3 billion parameters ~\citep{chung2022scaling}. Appendix \ref{supp-implementation} details our prompts and implementation.

\section{Experiments}
\label{sec:exp}
We demonstrate the effectiveness of \proposedmethod on two personalized text generation datasets from social media platforms.
For evaluation, we employ standard intrinsic evaluations, extrinsic evaluation based on downstream tasks using the generated text, and recently proposed personalized LLM-as-judge \cite{wang2023automated}. Then, in \S\ref{sec-routing-results} we show how a calibrated retriever can be used for selective revision of underperforming requests. We present ablations in \S\ref{sec-ablations} and we demonstrate the calibration performance for our retriever in \S\ref{sec-calibration-eval}.

\subsection{Experimental Setup}
\label{exp:data}
\textbf{Data} 
For evaluation, we use two open-ended long-form text generation datasets for social media: (1) Personalized post writing on \worksm and (2) Personalized comment writing on \aita.

\ul{\textsc{WorkSM}} \worksm is an enterprise social network used for communication within organizations presenting a highly realistic platform for writing assistance. We obtain a random sample of $\sim$18k posts written by 1116 users from November 2020 to July 2023. To create an \emph{evaluation set}, we manually examine posts greater than 50 words and receiving $\geq$2 comments, about 1K posts, and select 163 of the most recent posts from $\sim$80 users to serve as reference target texts $t_u^*$.
These posts represent a diverse, engaging set that could benefit from personalized writing assistance and serve as high quality target references. At a high level, these posts  share events, research studies, campaigns, and organizational news. Since \worksm does not contain requests to the writing assistant, two authors not involved in model development manually wrote requests $q_u$ per target text. Note that this was necessary given the highly regulated and private enterprise data in \worksm preventing exposure to external crowdworkers.  Our requests were authored following Guideline~\ref{fig-req-guidelines}. 
To construct $\historicalposts$ posts created before $t_u^*$ were used:
On average, users had $31$ historic posts (max of $169$). To create our \emph{training set}, we only retain posts $>10$ words and users with $\geq 5$ historic posts while excluding posts in our evaluation set. We generate synthetic requests with GPT-4 for training given the expense of manually authored requests -- resulting in a set of $\sim$7k training requests. Enterprise contracts with API providers ensured the privacy of user data shared over the API.

\ul{AITA} \aita is a Reddit subforum in which original posters (OP) describe personal moral conflicts and receive comments from other users judging them to be ``the a**hole'' or ``not the a**hole''. This dataset has been used in prior work on modeling the personal values of users \cite{plepi2022unifying}. We construct a personalized comment generation task from this data. 
We treat the OP posts as requests $q_u$, user comments as reference target texts $t_u^*$, and a user's previous comments as  $\historicalposts$. Since the dataset lacks time metadata, we construct an \emph{evaluation set} by sampling 10\% of the posts as test requests, and further filter to $600$ random target texts for our evaluation set to keep LLM experiments feasible. Evaluation users had $29$ posts in $\historicalposts$ on average (max of $590$). Our \emph{training set} used the historical post-comment pairs from users in $\historicalposts$, resulting in $\sim$84k requests. Note that while Reddit comments are not the ideal platform for writing assistance, \aita is one of the few public datasets available for the task and resembles applications such as email response generation \citep{kannan2016spartreply}. Appendix \ref{supp-datasets} details our datasets further. 
\begin{table*}[]
\subfloat[Extrinsic classification accuracy in \aita. \label{tab-aita-main-results}]{
\scalebox{0.71}{
\begin{tabular}{rcc}
\toprule
LLM $\rightarrow$ & \texttt{davinci-003} & \texttt{gpt-35-turbo}\\
\cmidrule(lr){2-2}\cmidrule(lr){3-3}
Method $\downarrow$ & {Macro F1(\%)} & {Macro F1(\%)}\\
\toprule
\textsc{zShot-NP} & 41.97 & 50.43\\
\textsc{kShot-NP} & 51.71 & 59.76\\
\midrule
Random & 55.52 & 59.47\\
BM25 & \underline{57.26} & \underline{61.66} \\
\textsc{MPNet-1B} & 53.72 & 59.23\\
UPR & 55.76 & 58.15\\
RelevanceCE &  56.85 & 59.59\\
\midrule
\textbf{\proposedmethod} & \textbf{\underline{61.21}} & \textbf{\underline{65.34}}\\
\bottomrule
\end{tabular}}}
\quad
\subfloat[Intrinsic reference based metrics in \worksm \label{tab-worksm-main-results}.]{
\centering
\scalebox{0.71}{
\begin{tabular}{rcccccc}
\toprule
~LLM $\rightarrow$ & \multicolumn{3}{c}{\texttt{davinci-003}} & \multicolumn{3}{c}{\texttt{gpt-35-turbo}}\\
\cmidrule(lr){2-4}\cmidrule(lr){5-7}
Method $\downarrow$~ & \small{\textsc{BS-F1}} & \small{\textsc{R1}} & \small{\textsc{R2}} & \small{\textsc{BS-F1}} & \small{\textsc{R1}} & \small{\textsc{R2}}\\
\toprule
\textsc{zShot-NP} & 36.25 & 0.5029	& 0.2516 & 31.03 & 0.4627 & 0.2091\\
\textsc{kShot-NP} & 34.08 & 0.4931 & 0.2431 & 32.51 & 0.4825 & 0.2258\\
\midrule
Random & 35.04 & 0.5036 & 0.2505 & 33.46 & 0.4893 & 0.2345\\
BM25 & 37.96 & 0.5287 & 0.2911 & \textbf{\underline{36.57}} & \textbf{\underline{0.5089}} & \underline{0.2673}\\
\textsc{MPNet-1B} & 38.30 & 0.5281 & 0.2931 & 36.02 & 0.5063 & 0.2639\\
UPR & \underline{38.70} & \underline{0.5337} & \underline{0.3019} & 35.98 & 0.5054 & 0.2642\\
RelevanceCE & 37.81 & 0.5288 & 0.2953 & 35.99 & 0.5038 & 0.2613\\
\midrule
\textbf{\proposedmethod} & \textbf{\underline{39.60}} & \textbf{\underline{0.5419}} & \textbf{\underline{0.3094}} & \underline{36.49} & \underline{0.5082} & \textbf{\underline{0.2676}}\\
\bottomrule
\end{tabular}}}
\caption{\proposedmethod is compared to non-personalized (NP) and LLMs personalized with retrieval on datasets of social media communication: (a) a dataset constructed from Reddit and (b) a workplace social media dataset.} 
\label{tab-both-main-results}
\end{table*}

\textbf{Generation metrics}
Since personalized text generation aims to adhere to the style, knowledge, and values of \emph{specific} users, effective evaluation for personalized generation remains an open problem \cite{wang2023automated, wang2023personalizedeval}. This is in contrast to non-personalized generation, where desirable aspects of outputs can be defined uniformly across all test cases. 
As a result, we present evaluations using a host of standard evaluation setups aiming to demonstrate the effectiveness of \proposedmethod from various perspectives. Our evaluations span the following standard setups \cite{dou2023automatic}: intrinsic evaluations based on n-gram/embedding similarity to reference texts, extrinsic evaluation through a classification accuracy based on generated text, and pairwise evaluation with personalized LLM-as-judge.

Specifically, for \worksm we report standard evaluation measures based on n-gram and embedding similarity between generations and reference targets: {\small ROUGE-1} (R1), {\small ROUGE-2} (R2), and BertScore-F1 (BS-F1) \citep{Zhang2020BERTScore}. This serves as an intrinsic evaluation for \worksm measuring the extent to which generations are similar to user authored texts. Next, since \aita users' comments primarily make a stance based on users' moral values, we measure if the stance in generated comments matches that of the user through a downstream {stance prediction} task -- serving as an extrinsic evaluation. This evaluation may be seen as evaluating the extent to which model generations adhere to a user's values.
We map generated comments to a binary ``YTA'' or ``NTA'' label based on simple high-precision rules mapping lexical variations of ``you're the a**hole'' and ``not the a**hole'' to the labels. This procedure was also found reliable for constructing ground truth labels in \aita \citep{plepi2022unifying}. Note that early attempts of using n-gram/embedding similarity measures for evaluation  (BS-F1, R1, R2) resulted in unreliable evaluations for \aita due the large variation (length, vocabulary, emojis etc.) in \aita comments, therefore we opt for more stable extrinsic evaluations and LLM based evaluations described next.

For both \aita and \worksm we conduct a pairwise evaluation with a recently proposed personalized LLM-as-judge \cite{wang2023automated}. Wang et al.\ show LLM based author identifications to be a reliable proxy task for distinguishing models of various qualities and being correlated with human quality ratings. Here, a judge LLM is presented with a reference text from a user and generations from the pair of systems being compared, then, it is prompted to select the system generation more likely to be authored by the author of the reference text. An author identification task aims to capture several aspects which distinguish individuals' writing, spanning style, knowledge and their values. In our evaluation, we compare \proposedmethod outputs to the outputs from the best baseline as indicated by intrinsic/extrinsic evaluations and use the target reference text $t_u^*$ in the LLM prompt as an example of the users writing. We use GPT-4o as our judge LLM and present the judge prompt in Appendix \ref{sec-exp-judgellmprompt}. In our evaluation we avoid rating aspects such as fluency, non-redundancy, etc.\ \citep{celikyilmaz2021evaluation} since we are primarily concerned with personalization performance and these qualities may be in conflict with specific users writing.

\textbf{Baselines}
As baselines, we consider non-personalized models based on zero shot prompting (\ul{\textsc{zShot-NP}}) and few-shot prompting with $k$ randomly chosen example documents (\ul{\textsc{kShot-NP}}). We consider retrieval-augmented personalized baselines, which selecting from a user's historical documents $\historicalposts$. They span selection at random from $\historicalposts$ (\ul{Random}), with sparse retrieval by \ul{BM25}, with dense retrieval by a strong \textsc{MPNet} model trained on 1 billion text pairs (\ul{\textsc{MPNet-1B}}), an unsupervised crossencoder \citep{sachan2022upr} ranking documents with \textsc{FlanT5-Base} likelihood: $p(\request | \doc)$ (\ul{UPR}), and a supervised crossencoder optimized on our dataset with request-document pairs, $(\request, \doc)$ in $\historicalposts$ (\ul{RelevanceCE}). Appendix \ref{supp-baselines} details our baselines.

\subsection{Generation Evaluation}
\label{sec-main-results}
Table \ref{tab-both-main-results} and \ref{tab-llmpair-eval} report our evaluations. Appendix \ref{supp-additional-res} presents ablation (\ref{sec-ablations}) and calibration (\ref{sec-calibration-eval}) results.

\textbf{Reference based evaluation}
Tables~\ref{tab-worksm-main-results} and \ref{tab-aita-main-results} reports automated metrics on \aita and \worksm. 
First we observe that personalization through retrieval, even at Random, generally improves upon non-personalized approaches (NP), which is consistent with prior work~\citep{salemi2023lamp}. Next, we note that the best baseline is not consistent, varying between BM25, and unsupervised crossencoder (UPR) -- indicating that retrieval models designed for request-document relevance vary in performance depending on the dataset and inference LLM. Finally, we note that \proposedmethod consistently performs at par or better than the best baselines across datasets and LLMs, indicating the effectiveness of training $\retriever$ for personalized generation.
For the more reliable classification metrics obtainable in \aita, \proposedmethod outperforms all baselines with improvements of 1.5 to 5 Macro F1 points. Next, we report performance in more expressive LLM-as-judge evaluations.
\begin{table}[]
\centering
\scalebox{0.8}{
\begin{tabular}{rcc}
\toprule
& \texttt{davinci-003} & \texttt{gpt-35-turbo}\\
\cmidrule(lr){2-2} \cmidrule(lr){3-3}
& P~/~B~/~T (\%) & P~/~B~/~T (\%)\\
\midrule
\aita & \textbf{46.8}~/40.3~/12.8$_{\alpha=0.56}$ & \textbf{46.6}~/44.9~/8.3$_{\alpha=0.55}$\\
\worksm & \textbf{46.6}~/42.5~/10.8$_{\alpha=0.42}$ & 38.9~/\textbf{42.6}~/18.5$_{\alpha=0.28}$\\
 \bottomrule
\end{tabular}}
\caption{LLM-as-judge win-rate evaluation for \aita and \worksm selecting a generation to be more aligned with an authors writing sample. The LLM could prefer the Proposed system (\proposedmethod), the Baseline (BM25), or judge the outputs as Tied -- denoted with P, B, and T.}
\label{tab-llmpair-eval}
\end{table}

\textbf{Pairwise LLM-as-judge evaluation}
In Table \ref{tab-llmpair-eval} we report the results of personalization evaluation following the setup described in \S\ref{exp:data}. Here, we compare against BM25-augmented as it performs within our top 2 baselines in automatic evaluations  - this strong performance is consistent with prior work \citep{izacard2022unsupervised, thakur2021beir}. We use GPT-4o as a judge LLM and run every pair of inputs through the judge LLM 3 times, we report average win rates over all the instances in our test set and over 3 repeated runs. Further, we randomly swap the position of the baseline and proposed method generations in the prompt to account for position biases in the judge LLM. Finally, we also report the agreement between the 3 judge LLM runs using Krippendorff's alpha ($\alpha$) to ensure that LLM judgements are consistent across runs.

In Table \ref{tab-llmpair-eval}, \proposedmethod achieves a greater win-rate than BM25 in 3 of 4 settings. In these settings we also note that the LLM judgments remain consistent across 3 repeated runs with Krippendorff's alpha between $0.41-0.56$ (0 indicates chance agreement). While BM25 sees a greater win-rate in \worksm with \texttt{gpt-35-turbo}, the judgments see lower agreement ($\alpha=0.28$) indicating the outputs to be harder to distinguish. Finally, comparing to Table \ref{tab-both-main-results} we see that the trends of extrinsic and intrinsic reference based evaluations are retained in LLM-as-judge evaluations -- consistently indicating the benefit of \proposedmethod across evaluation setups, inference LLMs, and datasets. In Appendix \ref{supp-additional-res} we show an example from \aita to show the kinds of retrievals and outputs that make \proposedmethod effective.

\subsection{Selective Revision with \proposedmethod}
\label{sec-routing-results}
\begin{figure}[t]
     \centering
     {\includegraphics[width=0.7\textwidth]{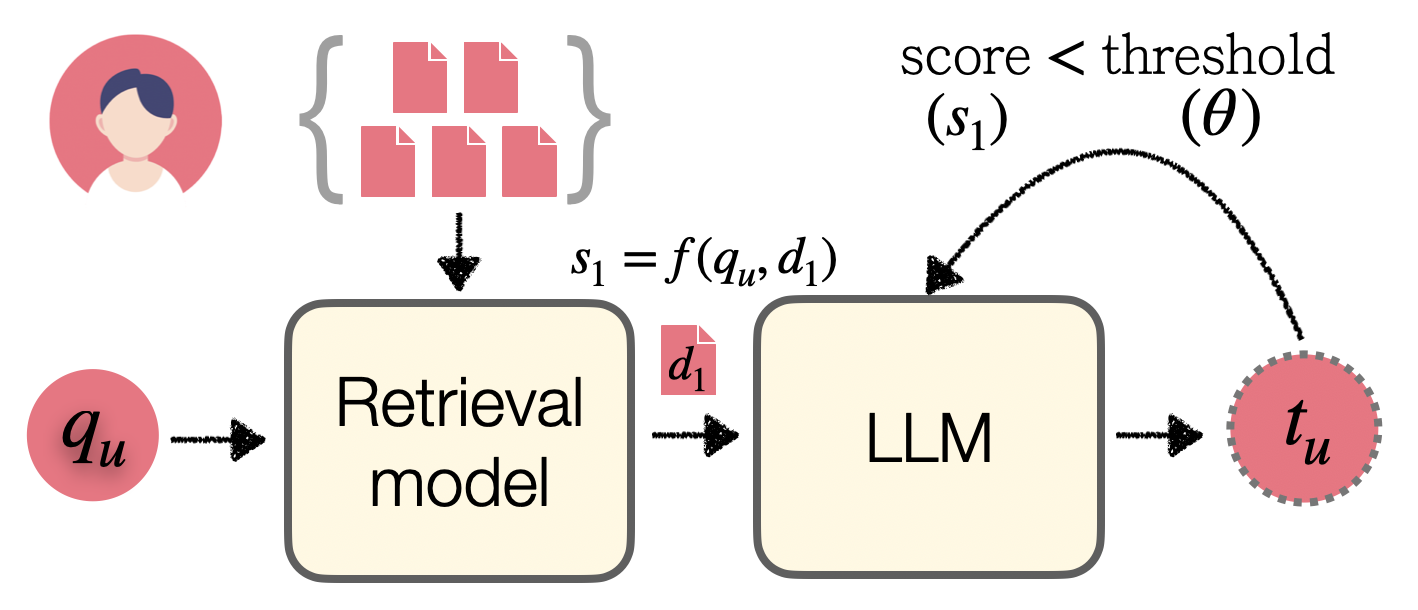}}
     \caption{Generation calibration of $\retriever$ allows us to use its predicted scores for performance prediction and selectively revise potentially bad generations.}
     \label{fig-revision}
 \end{figure}

Having established \proposedmethod to be an effective model for generation, we show $\retriever$ to be generation calibrated in Appendix \ref{sec-calibration-eval}. Here, we demonstrate the usefulness of a calibrated retriever in a case study using the retriever scores to selectively revise generations.
Specifically, we treat the scores from $\retriever$ as a {predictor of retrieval performance, and in-turn text generation performance}.
We assume that if $\retriever$ cannot find a highly scored in-context example, the generated response will be of low quality and can benefit from LLM revision (Figure \ref{fig-revision}).

\textbf{Setup}
Given our trained retriever, we take all top-1 document scores for each request $s_1 = \max_{\doc \in \historicalposts} \retriever(\request, \doc)$
and learn a threshold $\theta$ on $s_1$ that maximizes a downstream performance metric on a held-out development set ({\small R2} in \worksm and Macro-F1 in \aita). Then, given a generated target text $\target$ with $s_1 < \theta$, we selectively revise $\target$ where $\llmG$ is prompted to edit the target text. We report results of selective revision compared to a single round of generation (i.e., no revision) and full revision over the entire dataset (i.e., 100\% revision). We repeat this for BM25.
We provide further details and analysis in Appendix \ref{supp-selective-editing}.
\begin{table}[]
\centering
\scalebox{0.71}{
\begin{tabular}{rcccc}
\toprule
Dataset $\rightarrow$ & \multicolumn{1}{c}{\aita} & \multicolumn{3}{c}{\worksm}\\
\cmidrule(lr){2-2}\cmidrule(lr){3-5}
Method $\downarrow$~/~LLM $\rightarrow$ & \multicolumn{1}{c}{\texttt{gpt-35-turbo}} & \multicolumn{3}{c}{\texttt{gpt-35-turbo}}\\
\cmidrule(lr){2-2}\cmidrule(lr){3-5}
$\retriever=$~BM25  & \small{Macro F1 (\%)} & \small{\textsc{BS-F1}} & \small{\textsc{R1}} & \small{\textsc{R2}} \\
\toprule
Stage 1 (no revision) & \textbf{59.99} & 36.15 & 0.5052 & 0.2611 \\
All (100\% revision) & 58.36 & 35.45 & 0.5096 & 0.2573 \\
Selective revision & 57.71 & \textbf{37.29} & \textbf{0.5206} & \textbf{0.2738}\\
\bottomrule
& \\
\midrule
$\retriever=$~Proposed  & \small{Macro F1 (\%)} & \small{\textsc{BS-F1}} & \small{\textsc{R1}} & \small{\textsc{R2}}\\
\toprule
 Stage 1 (no revision) & 65.15 & 37.02 & 0.5124 & 0.2709\\
 All (100\% revision) & 64.85 & 35.47 & 0.5045 & 0.2520\\
 Selective revision & \textbf{65.36} & \textbf{37.71} & \textbf{0.5236} & \textbf{0.2818}\\
\bottomrule
\end{tabular}}
\caption{Selectively revising target texts $t_u$ based on scores from our retriever vs BM25. Also present are results of no revision and revising all outputs (100\% revision) from Stage 1 outputs.}
\label{tab-both-routing-baseline}
\end{table}

\textbf{Results}
In Table \ref{tab-both-routing-baseline} we see that 
selective revision improves or retains performance upon a single round of generation (``Stage 1'') by 2-4\% in downstream performance metrics with $\retriever=$Proposed and BM25 for \worksm. However, for \aita we see that selective revision based on BM25 shows a marked drop in performance indicating its dataset dependent calibration performance. Importantly, note that Macro F1 doesn't measure aspects of style which may have changed in revision. Finally, editing \textit{all} outputs produced by Stage 1 generation consistently leads to degraded performance (``All''), indicating that editing is not always helpful.

We also observe that \proposedmethod chooses $75.8\%$ and $77.9\%$ instances for editing in \worksm and \aita, respectively. This indicates the potential for generation calibrated retrievers to reduce the number of expensive LLM calls made while ensuring better personalization performance. In Figure \ref{fig-selective-rev-perq} (Appendix \ref{supp-selective-editing}) we analyze the performance of selective revision against request and user profile length. In a manual examination of requests with a low $s_1$ score by the \proposedmethod $\retriever$, we find the requests to be underspecified and often require further information from a user e.g. the request ``Write a post about how I like to relax after work'', aims to generate a target discussing more specific forms of relaxation not present in any historical documents. This indicates that generation calibrated retrievers may be used for other forms of selective prediction and user interaction -- e.g.\ selectively withholding predictions when satisfactory generations are unlikely or obtaining more information from users through follow-up questions. We leave such explorations to exciting future work.

\section{Conclusion}
\label{sec:conclusion}
In this paper we present \proposedmethod -- an  LLM based writing assistant personalized with generation calibrated retrievers. We propose a method for training generation calibrated retrievers through a careful selection of training data and a scale calibrated objective. In a series of holistic evaluations, we demonstrate the effectiveness of our approach in datasets of social media communication compared to baselines (\S\ref{sec-main-results}) as well as ablated models (Appendix \ref{sec-ablations}). We demonstrate the calibration performance for our retriever (Appendix \ref{sec-calibration-eval}), and show how our retrieval model can double as a performance predictor (\S\ref{sec-routing-results}) and can identify outputs which can benefit from LLM revision.

\section{Ethical and broader impact}
\label{sec-eth-broad-impact}
Having introduced \proposedmethod as an effective personalization strategy for writing assistance and discussed its benefits we review two  implications of concern arising from better personalized text generation: challenges to factuality, and longer term influence on language use and communication.

\textbf{Challenges to factuality} The emergence of LLMs and their ability to generate compelling text has seen a subsequent rise in the cases of malicious use of these technologies.
\citet{augenstein2023factuality} overview four such classes of harm: personalized attacks on individuals in the form of phishing attacks and tailored misinformation, impersonation of trusted figures (e.g.\ journalists or regulators), a glut of paraphrased misinformation evading detection by automatic tools often used by fact checkers, and large scale creation of fake social media profiles and plagiarized content \citep{brewster2023plagiarism}. It is possible that improvements in personalized text generation are likely to excacertabe each of these problems. To account for this, several technology and policy initiatives are under active development \cite{augenstein2023factuality}. These span detection of AI-generated content, cryptographic signatures intended to prove the authenticity of content, to government regulation and public education, however, their effectiveness remains under investigation.

\textbf{Language use and communication} Current understanding of computer mediated communication suggests that users interpersonal communication patterns are influenced by the tool/medium used for communication \citep{poddar2023topicchange} with a potential for these influences to have longer term influences on  communication in the absence of these tools \citep{hancock2020aimc}. Hancock et al.\ outline these implications as ranging from shifts in language use (e.g\ a social expectation of more positive responses \citep{hohenstein2018allopositive}), issues of how individuals portray themselves and evaluate others, to long term feedback loops resulting in how we perceive ourselves. However, understanding of the implications of AI mediated communication, specially those powered by powerful LLMs, is largely developing \citep{hancock2020aimc}. It is likely that wide spread personalization in LLM communication agents, will necessitate further understanding of these factors and the design of systems that incorporates this understanding to ameliorate harms.

\bibliography{generation_cal_retriever}

\begin{thebibliography}{78}
\expandafter\ifx\csname natexlab\endcsname\relax\def\natexlab#1{#1}\fi

\bibitem[{Akoury et~al.(2020)Akoury, Wang, Whiting, Hood, Peng, and
  Iyyer}]{akoury2020storium}
Nader Akoury, Shufan Wang, Josh Whiting, Stephen Hood, Nanyun Peng, and Mohit
  Iyyer. 2020.
\newblock \href {https://doi.org/10.18653/v1/2020.emnlp-main.525} {{STORIUM}:
  {A} {D}ataset and {E}valuation {P}latform for {M}achine-in-the-{L}oop {S}tory
  {G}eneration}.
\newblock In \emph{Proceedings of the 2020 Conference on Empirical Methods in
  Natural Language Processing (EMNLP)}, pages 6470--6484, Online. Association
  for Computational Linguistics.

\bibitem[{Ao et~al.(2021)Ao, Wang, Luo, Qiao, He, and Xie}]{ao2021pens}
Xiang Ao, Xiting Wang, Ling Luo, Ying Qiao, Qing He, and Xing Xie. 2021.
\newblock \href {https://doi.org/10.18653/v1/2021.acl-long.7} {{PENS}: A
  dataset and generic framework for personalized news headline generation}.
\newblock In \emph{Proceedings of the 59th Annual Meeting of the Association
  for Computational Linguistics and the 11th International Joint Conference on
  Natural Language Processing (Volume 1: Long Papers)}, pages 82--92, Online.
  Association for Computational Linguistics.

\bibitem[{Arora et~al.(2023)Arora, Narayan, Chen, Orr, Guha, Bhatia, Chami, and
  Re}]{arora2023ask}
Simran Arora, Avanika Narayan, Mayee~F Chen, Laurel Orr, Neel Guha, Kush
  Bhatia, Ines Chami, and Christopher Re. 2023.
\newblock \href {https://openreview.net/forum?id=bhUPJnS2g0X} {Ask me anything:
  A simple strategy for prompting language models}.
\newblock In \emph{The Eleventh International Conference on Learning
  Representations}.

\bibitem[{Augenstein et~al.(2023)Augenstein, Baldwin, Cha, Chakraborty,
  Ciampaglia, Corney, DiResta, Ferrara, Hale, Halevy, Hovy, Ji, Menczer,
  Miguez, Nakov, Scheufele, Sharma, and Zagni}]{augenstein2023factuality}
Isabelle Augenstein, Timothy Baldwin, Meeyoung Cha, Tanmoy Chakraborty,
  Giovanni~Luca Ciampaglia, David Corney, Renee DiResta, Emilio Ferrara, Scott
  Hale, Alon Halevy, Eduard Hovy, Heng Ji, Filippo Menczer, Ruben Miguez,
  Preslav Nakov, Dietram Scheufele, Shivam Sharma, and Giovanni Zagni. 2023.
\newblock \href {http://arxiv.org/abs/2310.05189} {Factuality challenges in the
  era of large language models}.
\newblock \emph{arXiv preprint arXiv:2310.05189}.

\bibitem[{August et~al.(2022)August, Reinecke, and
  Smith}]{august2022generating}
Tal August, Katharina Reinecke, and Noah~A. Smith. 2022.
\newblock \href {https://doi.org/10.18653/v1/2022.acl-long.569} {Generating
  scientific definitions with controllable complexity}.
\newblock In \emph{Proceedings of the 60th Annual Meeting of the Association
  for Computational Linguistics (Volume 1: Long Papers)}, pages 8298--8317,
  Dublin, Ireland. Association for Computational Linguistics.

\bibitem[{Bonifacio et~al.(2022)Bonifacio, Abonizio, Fadaee, and
  Nogueira}]{bonifacio2022inpars}
Luiz Bonifacio, Hugo Abonizio, Marzieh Fadaee, and Rodrigo Nogueira. 2022.
\newblock \href {https://doi.org/10.1145/3477495.3531863} {Inpars: Unsupervised
  dataset generation for information retrieval}.
\newblock In \emph{Proceedings of the 45th International ACM SIGIR Conference
  on Research and Development in Information Retrieval}, SIGIR '22, page
  2387–2392, New York, NY, USA. Association for Computing Machinery.

\bibitem[{Brewster et~al.(2023)Brewster, Wang, and
  Palmer}]{brewster2023plagiarism}
Jack Brewster, Macrina Wang, and Coalter Palmer. 2023.
\newblock \href
  {https://www.newsguardtech.com/misinformation-monitor/august-2023/}
  {Plagiarism-bot? how low-quality websites are using ai to deceptively rewrite
  content from mainstream news outlets}.
\newblock NewsGaurd, The Internet Trust Tool.

\bibitem[{Celikyilmaz et~al.(2021)Celikyilmaz, Clark, and
  Gao}]{celikyilmaz2021evaluation}
Asli Celikyilmaz, Elizabeth Clark, and Jianfeng Gao. 2021.
\newblock \href {http://arxiv.org/abs/2006.14799} {Evaluation of text
  generation: A survey}.
\newblock \emph{arXiv preprint arXiv:2006.14799}.

\bibitem[{Chakrabarty et~al.(2022)Chakrabarty, Padmakumar, and
  He}]{chakrabarty2022instpoem}
Tuhin Chakrabarty, Vishakh Padmakumar, and He~He. 2022.
\newblock \href {https://doi.org/10.18653/v1/2022.emnlp-main.460} {Help me
  write a poem: Instruction tuning as a vehicle for collaborative poetry
  writing}.
\newblock In \emph{Proceedings of the 2022 Conference on Empirical Methods in
  Natural Language Processing}, pages 6848--6863, Abu Dhabi, United Arab
  Emirates. Association for Computational Linguistics.

\bibitem[{Chen et~al.(2019)Chen, Lee, Bansal, Cao, Zhang, Lu, Tsay, Wang, Dai,
  Chen, Sohn, and Wu}]{chen2019gmail}
Mia~Xu Chen, Benjamin~N. Lee, Gagan Bansal, Yuan Cao, Shuyuan Zhang, Justin Lu,
  Jackie Tsay, Yinan Wang, Andrew~M. Dai, Zhifeng Chen, Timothy Sohn, and
  Yonghui Wu. 2019.
\newblock \href {https://doi.org/10.1145/3292500.3330723} {Gmail smart compose:
  Real-time assisted writing}.
\newblock In \emph{Proceedings of the 25th ACM SIGKDD International Conference
  on Knowledge Discovery \& Data Mining}, KDD '19, page 2287–2295, New York,
  NY, USA. Association for Computing Machinery.

\bibitem[{Chen et~al.(2023)Chen, Brumby, and Cox}]{chen2023peerwrite}
Shiping Chen, Duncan~P Brumby, and Anna~L Cox. 2023.
\newblock \href
  {https://cdn.glitch.global/d058c114-3406-43be-8a3c-d3afff35eda2/paper37_2023.pdf}
  {Using writing assistants to accelerate the peer review process}.
\newblock \emph{Second Workshop on Intelligent and Interactive Writing
  Assistants, CHI 2023}.

\bibitem[{Chung et~al.(2022)Chung, Hou, Longpre, Zoph, Tay, Fedus, Li, Wang,
  Dehghani, Brahma et~al.}]{chung2022scaling}
Hyung~Won Chung, Le~Hou, Shayne Longpre, Barret Zoph, Yi~Tay, William Fedus,
  Eric Li, Xuezhi Wang, Mostafa Dehghani, Siddhartha Brahma, et~al. 2022.
\newblock Scaling instruction-finetuned language models.
\newblock \emph{arXiv preprint arXiv:2210.11416}.

\bibitem[{Clark et~al.(2018)Clark, Ross, Tan, Ji, and Smith}]{clark2018poems}
Elizabeth Clark, Anne~Spencer Ross, Chenhao Tan, Yangfeng Ji, and Noah~A.
  Smith. 2018.
\newblock \href {https://doi.org/10.1145/3172944.3172983} {Creative writing
  with a machine in the loop: Case studies on slogans and stories}.
\newblock In \emph{23rd International Conference on Intelligent User
  Interfaces}, IUI '18, page 329–340, New York, NY, USA. Association for
  Computing Machinery.

\bibitem[{Cohen et~al.(2021)Cohen, Mitra, Lesota, Rekabsaz, and
  Eickhoff}]{cohen2021uncertainrel}
Daniel Cohen, Bhaskar Mitra, Oleg Lesota, Navid Rekabsaz, and Carsten Eickhoff.
  2021.
\newblock \href {https://doi.org/10.1145/3404835.3462951} {Not all relevance
  scores are equal: Efficient uncertainty and calibration modeling for deep
  retrieval models}.
\newblock In \emph{Proceedings of the 44th International ACM SIGIR Conference
  on Research and Development in Information Retrieval}, SIGIR '21, page
  654–664, New York, NY, USA. Association for Computing Machinery.

\bibitem[{Creswell et~al.(2023)Creswell, Shanahan, and
  Higgins}]{creswell2023selectioninference}
Antonia Creswell, Murray Shanahan, and Irina Higgins. 2023.
\newblock \href {https://openreview.net/forum?id=3Pf3Wg6o-A4}
  {Selection-inference: Exploiting large language models for interpretable
  logical reasoning}.
\newblock In \emph{The Eleventh International Conference on Learning
  Representations}.

\bibitem[{Dale and Viethen(2021)}]{dale_viethen_2021}
Robert Dale and Jette Viethen. 2021.
\newblock \href {https://doi.org/10.1017/S1351324921000164} {The automated
  writing assistance landscape in 2021}.
\newblock \emph{Natural Language Engineering}, 27(4):511–518.

\bibitem[{Dhuliawala et~al.(2022)Dhuliawala, Adolphs, Das, and
  Sachan}]{dhuliawala2022mrcalibration}
Shehzaad Dhuliawala, Leonard Adolphs, Rajarshi Das, and Mrinmaya Sachan. 2022.
\newblock \href {https://doi.org/10.18653/v1/2022.findings-acl.133}
  {Calibration of machine reading systems at scale}.
\newblock In \emph{Findings of the Association for Computational Linguistics:
  ACL 2022}, pages 1682--1693, Dublin, Ireland. Association for Computational
  Linguistics.

\bibitem[{Dhuliawala et~al.(2023)Dhuliawala, Komeili, Xu, Raileanu, Li,
  Celikyilmaz, and Weston}]{dhuliawala2023chainofverification}
Shehzaad Dhuliawala, Mojtaba Komeili, Jing Xu, Roberta Raileanu, Xian Li, Asli
  Celikyilmaz, and Jason Weston. 2023.
\newblock \href {http://arxiv.org/abs/2309.11495} {Chain-of-verification
  reduces hallucination in large language models}.
\newblock \emph{arXiv preprint arXiv:2309.11495}.

\bibitem[{Dou et~al.(2023)Dou, Laban, Gardent, and Xu}]{dou2023automatic}
Yao Dou, Philippe Laban, Claire Gardent, and Wei Xu. 2023.
\newblock Automatic and human-ai interactive text generation.
\newblock \emph{arXiv preprint arXiv:2310.03878}.

\bibitem[{Farajidizaji et~al.(2023)Farajidizaji, Raina, and
  Gales}]{farajidizaji2023possible}
Asma Farajidizaji, Vatsal Raina, and Mark Gales. 2023.
\newblock \href {http://arxiv.org/abs/2309.12551} {Is it possible to modify
  text to a target readability level? an initial investigation using zero-shot
  large language models}.
\newblock \emph{arXiv preprint arXiv:2309.12551}.

\bibitem[{Gero et~al.(2022)Gero, Liu, and Chilton}]{gero2022sparks}
Katy~Ilonka Gero, Vivian Liu, and Lydia Chilton. 2022.
\newblock \href {https://doi.org/10.1145/3532106.3533533} {Sparks: Inspiration
  for science writing using language models}.
\newblock In \emph{Proceedings of the 2022 ACM Designing Interactive Systems
  Conference}, DIS '22, page 1002–1019, New York, NY, USA. Association for
  Computing Machinery.

\bibitem[{Gon{\c{c}}alo~Oliveira(2017)}]{goncalo2017poemsurvey}
Hugo Gon{\c{c}}alo~Oliveira. 2017.
\newblock \href {https://doi.org/10.18653/v1/W17-3502} {A survey on intelligent
  poetry generation: Languages, features, techniques, reutilisation and
  evaluation}.
\newblock In \emph{Proceedings of the 10th International Conference on Natural
  Language Generation}, pages 11--20, Santiago de Compostela, Spain.
  Association for Computational Linguistics.

\bibitem[{Gonen et~al.(2022)Gonen, Iyer, Blevins, Smith, and
  Zettlemoyer}]{gonen2022demystifying}
Hila Gonen, Srini Iyer, Terra Blevins, Noah~A. Smith, and Luke Zettlemoyer.
  2022.
\newblock \href {http://arxiv.org/abs/2212.04037} {Demystifying prompts in
  language models via perplexity estimation}.
\newblock \emph{arXiv preprint arXiv:2212.04037}.

\bibitem[{Guo et~al.(2021)Guo, Qiu, Wang, and Cohen}]{guo2021laysumm}
Yue Guo, Wei Qiu, Yizhong Wang, and Trevor Cohen. 2021.
\newblock \href {https://doi.org/10.1609/aaai.v35i1.16089} {Automated lay
  language summarization of biomedical scientific reviews}.
\newblock \emph{Proceedings of the AAAI Conference on Artificial Intelligence},
  35(1):160--168.

\bibitem[{Gupta et~al.(2023)Gupta, Gardner, and Singh}]{gupta2023coveragebased}
Shivanshu Gupta, Matt Gardner, and Sameer Singh. 2023.
\newblock \href {http://arxiv.org/abs/2305.14907} {Coverage-based example
  selection for in-context learning}.
\newblock \emph{arXiv preprint arXiv:2305.14907}.

\bibitem[{Hancock et~al.(2020)Hancock, Naaman, and Levy}]{hancock2020aimc}
Jeffrey~T Hancock, Mor Naaman, and Karen Levy. 2020.
\newblock \href {https://doi.org/10.1093/jcmc/zmz022} {{AI-Mediated
  Communication: Definition, Research Agenda, and Ethical Considerations}}.
\newblock \emph{Journal of Computer-Mediated Communication}, 25(1):89--100.

\bibitem[{Hohenstein and Jung(2018)}]{hohenstein2018allopositive}
Jess Hohenstein and Malte Jung. 2018.
\newblock \href {https://doi.org/10.1145/3170427.3188487} {Ai-supported
  messaging: An investigation of human-human text conversation with ai
  support}.
\newblock In \emph{Extended Abstracts of the 2018 CHI Conference on Human
  Factors in Computing Systems}, CHI EA '18, page 1–6, New York, NY, USA.
  Association for Computing Machinery.

\bibitem[{Hu(2023)}]{krystal2023chatgptrec}
Krystal Hu. 2023.
\newblock \href
  {https://www.reuters.com/technology/chatgpt-sets-record-fastest-growing-user-base-analyst-note-2023-02-01/}
  {Chatgpt sets record for fastest-growing user base - analyst note}.
\newblock National Bureau of Economic Research, Digest No. 6.

\bibitem[{Izacard et~al.(2022)Izacard, Caron, Hosseini, Riedel, Bojanowski,
  Joulin, and Grave}]{izacard2022unsupervised}
Gautier Izacard, Mathilde Caron, Lucas Hosseini, Sebastian Riedel, Piotr
  Bojanowski, Armand Joulin, and Edouard Grave. 2022.
\newblock \href {https://openreview.net/forum?id=jKN1pXi7b0} {Unsupervised
  dense information retrieval with contrastive learning}.
\newblock \emph{Transactions on Machine Learning Research}.

\bibitem[{Kannan et~al.(2016)Kannan, Kurach, Ravi, Kaufmann, Tomkins, Miklos,
  Corrado, Lukacs, Ganea, Young, and Ramavajjala}]{kannan2016spartreply}
Anjuli Kannan, Karol Kurach, Sujith Ravi, Tobias Kaufmann, Andrew Tomkins,
  Balint Miklos, Greg Corrado, Laszlo Lukacs, Marina Ganea, Peter Young, and
  Vivek Ramavajjala. 2016.
\newblock \href {https://doi.org/10.1145/2939672.2939801} {Smart reply:
  Automated response suggestion for email}.
\newblock In \emph{Proceedings of the 22nd ACM SIGKDD International Conference
  on Knowledge Discovery and Data Mining}, KDD '16, page 955–964, New York,
  NY, USA. Association for Computing Machinery.

\bibitem[{Khattab et~al.(2023)Khattab, Santhanam, Li, Hall, Liang, Potts, and
  Zaharia}]{khattab2023demonstratesearchpredict}
Omar Khattab, Keshav Santhanam, Xiang~Lisa Li, David Hall, Percy Liang,
  Christopher Potts, and Matei Zaharia. 2023.
\newblock \href {http://arxiv.org/abs/2212.14024} {Demonstrate-search-predict:
  Composing retrieval and language models for knowledge-intensive nlp}.
\newblock \emph{arXiv preprint arXiv:2212.14024}.

\bibitem[{Kirk et~al.(2023)Kirk, Vidgen, R{\"o}ttger, and
  Hale}]{kirk2023personalisation}
Hannah~Rose Kirk, Bertie Vidgen, Paul R{\"o}ttger, and Scott~A Hale. 2023.
\newblock Personalisation within bounds: A risk taxonomy and policy framework
  for the alignment of large language models with personalised feedback.
\newblock \emph{arXiv preprint arXiv:2303.05453}.

\bibitem[{Li et~al.(2023{\natexlab{a}})Li, Zhang, Mei, Kong, and
  Bendersky}]{li2023automatic}
Cheng Li, Mingyang Zhang, Qiaozhu Mei, Weize Kong, and Michael Bendersky.
  2023{\natexlab{a}}.
\newblock \href {http://arxiv.org/abs/2310.00152} {Automatic prompt rewriting
  for personalized text generation}.
\newblock \emph{arXiv preprint arXiv:2310.00152}.

\bibitem[{Li et~al.(2023{\natexlab{b}})Li, Zhang, Mei, Wang, Hombaiah, Liang,
  and Bendersky}]{li2023teach}
Cheng Li, Mingyang Zhang, Qiaozhu Mei, Yaqing Wang, Spurthi~Amba Hombaiah,
  Yi~Liang, and Michael Bendersky. 2023{\natexlab{b}}.
\newblock \href {http://arxiv.org/abs/2308.07968} {Teach llms to personalize --
  an approach inspired by writing education}.
\newblock \emph{arXiv preprint arXiv:2308.07968}.

\bibitem[{Li et~al.(2023{\natexlab{c}})Li, Rawat, Zaheer, Wang, Lukasik, Veit,
  Yu, and Kumar}]{li2023controlcontext}
Daliang Li, Ankit~Singh Rawat, Manzil Zaheer, Xin Wang, Michal Lukasik, Andreas
  Veit, Felix Yu, and Sanjiv Kumar. 2023{\natexlab{c}}.
\newblock \href {https://doi.org/10.18653/v1/2023.findings-acl.112} {Large
  language models with controllable working memory}.
\newblock In \emph{Findings of the Association for Computational Linguistics:
  ACL 2023}, pages 1774--1793, Toronto, Canada. Association for Computational
  Linguistics.

\bibitem[{Li et~al.(2023{\natexlab{d}})Li, Belkadi, Micheletti, Han, Shardlow,
  and Nenadic}]{li2023large}
Zihao Li, Samuel Belkadi, Nicolo Micheletti, Lifeng Han, Matthew Shardlow, and
  Goran Nenadic. 2023{\natexlab{d}}.
\newblock \href {http://arxiv.org/abs/2309.13202} {Large language models and
  control mechanisms improve text readability of biomedical abstracts}.

\bibitem[{Lin et~al.(2023)Lin, Ehsan, Agarwal, Dani, Vashishth, and
  Riedl}]{lin2023prompts}
Zhiyu Lin, Upol Ehsan, Rohan Agarwal, Samihan Dani, Vidushi Vashishth, and Mark
  Riedl. 2023.
\newblock \href
  {https://computationalcreativity.net/iccc23/papers/ICCC-2023_paper_25.pdf}
  {Beyond prompts: Exploring the design space of mixed-initiative co-creativity
  systems}.
\newblock In \emph{ICCC}.

\bibitem[{Luo et~al.(2023)Luo, Chuang, Gong, Zhang, Kim, Wu, Fox, Meng, and
  Glass}]{luo2023sail}
Hongyin Luo, Yung-Sung Chuang, Yuan Gong, Tianhua Zhang, Yoon Kim, Xixin Wu,
  Danny Fox, Helen Meng, and James Glass. 2023.
\newblock \href {http://arxiv.org/abs/2305.15225} {Sail: Search-augmented
  instruction learning}.
\newblock \emph{arXiv preprint arXiv:2305.15225}.

\bibitem[{Mahlow(2023)}]{mahlow2023writing}
Cerstin Mahlow. 2023.
\newblock \href {http://arxiv.org/abs/2303.17894} {Writing tools: Looking back
  to look ahead}.
\newblock \emph{Second Workshop on Intelligent and Interactive Writing
  Assistants, CHI 2023}.

\bibitem[{Majumder et~al.(2019)Majumder, Li, Ni, and
  McAuley}]{majumder2019generating}
Bodhisattwa~Prasad Majumder, Shuyang Li, Jianmo Ni, and Julian McAuley. 2019.
\newblock \href {https://doi.org/10.18653/v1/D19-1613} {Generating personalized
  recipes from historical user preferences}.
\newblock In \emph{Proceedings of the 2019 Conference on Empirical Methods in
  Natural Language Processing and the 9th International Joint Conference on
  Natural Language Processing (EMNLP-IJCNLP)}, pages 5976--5982, Hong Kong,
  China. Association for Computational Linguistics.

\bibitem[{Mallen et~al.(2023)Mallen, Asai, Zhong, Das, Khashabi, and
  Hajishirzi}]{mallen2023trustcontext}
Alex Mallen, Akari Asai, Victor Zhong, Rajarshi Das, Daniel Khashabi, and
  Hannaneh Hajishirzi. 2023.
\newblock \href {https://doi.org/10.18653/v1/2023.acl-long.546} {When not to
  trust language models: Investigating effectiveness of parametric and
  non-parametric memories}.
\newblock In \emph{Proceedings of the 61st Annual Meeting of the Association
  for Computational Linguistics (Volume 1: Long Papers)}, pages 9802--9822,
  Toronto, Canada. Association for Computational Linguistics.

\bibitem[{Mazar{\'e} et~al.(2018)Mazar{\'e}, Humeau, Raison, and
  Bordes}]{mazare2018training}
Pierre-Emmanuel Mazar{\'e}, Samuel Humeau, Martin Raison, and Antoine Bordes.
  2018.
\newblock \href {https://doi.org/10.18653/v1/D18-1298} {Training millions of
  personalized dialogue agents}.
\newblock In \emph{Proceedings of the 2018 Conference on Empirical Methods in
  Natural Language Processing}, pages 2775--2779, Brussels, Belgium.
  Association for Computational Linguistics.

\bibitem[{Menon et~al.(2022)Menon, Jayasumana, Rawat, Kim, Reddi, and
  Kumar}]{menon22defensedual}
Aditya Menon, Sadeep Jayasumana, Ankit~Singh Rawat, Seungyeon Kim, Sashank
  Reddi, and Sanjiv Kumar. 2022.
\newblock \href {https://proceedings.mlr.press/v162/menon22a.html} {In defense
  of dual-encoders for neural ranking}.
\newblock In \emph{Proceedings of the 39th International Conference on Machine
  Learning}, volume 162 of \emph{Proceedings of Machine Learning Research},
  pages 15376--15400. PMLR.

\bibitem[{Mirowski et~al.(2023)Mirowski, Mathewson, Pittman, and
  Evans}]{mirowski2023screens}
Piotr Mirowski, Kory~W. Mathewson, Jaylen Pittman, and Richard Evans. 2023.
\newblock \href {https://doi.org/10.1145/3544548.3581225} {Co-writing
  screenplays and theatre scripts with language models: Evaluation by industry
  professionals}.
\newblock In \emph{Proceedings of the 2023 CHI Conference on Human Factors in
  Computing Systems}, CHI '23, New York, NY, USA. Association for Computing
  Machinery.

\bibitem[{Murthy et~al.(2022)Murthy, Lo, King, Bhagavatula, Kuehl, Johnson,
  Borchardt, Weld, Hope, and Downey}]{murthy2022accord}
Sonia~K. Murthy, Kyle Lo, Daniel King, Chandra Bhagavatula, Bailey Kuehl,
  Sophie Johnson, Jonathan Borchardt, Daniel~S. Weld, Tom Hope, and Doug
  Downey. 2022.
\newblock \href {https://doi.org/10.48550/ARXIV.2205.06982} {Accord: A
  multi-document approach to generating diverse descriptions of scientific
  concepts}.

\bibitem[{Nadejde and Tetreault(2019)}]{nadejde2019personalgec}
Maria Nadejde and Joel Tetreault. 2019.
\newblock \href {https://doi.org/10.18653/v1/D19-5504} {Personalizing
  grammatical error correction: Adaptation to proficiency level and {L}1}.
\newblock In \emph{Proceedings of the 5th Workshop on Noisy User-generated Text
  (W-NUT 2019)}, pages 27--33, Hong Kong, China. Association for Computational
  Linguistics.

\bibitem[{Ni et~al.(2017)Ni, Lipton, Vikram, and McAuley}]{ni2017estimating}
Jianmo Ni, Zachary~C. Lipton, Sharad Vikram, and Julian McAuley. 2017.
\newblock \href {https://aclanthology.org/I17-1079} {Estimating reactions and
  recommending products with generative models of reviews}.
\newblock In \emph{Proceedings of the Eighth International Joint Conference on
  Natural Language Processing (Volume 1: Long Papers)}, pages 783--791, Taipei,
  Taiwan. Asian Federation of Natural Language Processing.

\bibitem[{Pan et~al.(2023)Pan, Yao, Zhang, Yu, Yu, and
  Chen}]{pan2023knowledgeselector}
Xiaoman Pan, Wenlin Yao, Hongming Zhang, Dian Yu, Dong Yu, and Jianshu Chen.
  2023.
\newblock \href {https://openreview.net/forum?id=a2jNdqE2102}
  {Knowledge-in-context: Towards knowledgeable semi-parametric language
  models}.
\newblock In \emph{The Eleventh International Conference on Learning
  Representations}.

\bibitem[{Papenmeier et~al.(2021)Papenmeier, Kern, Hienert, Sliwa, Aker, and
  Fuhr}]{papenmeier2021convsearch}
Andrea Papenmeier, Dagmar Kern, Daniel Hienert, Alfred Sliwa, Ahmet Aker, and
  Norbert Fuhr. 2021.
\newblock \href {https://doi.org/10.1145/3406522.3446035} {Starting
  conversations with search engines - interfaces that elicit natural language
  queries}.
\newblock In \emph{Proceedings of the 2021 Conference on Human Information
  Interaction and Retrieval}, CHIIR '21, page 261–265, New York, NY, USA.
  Association for Computing Machinery.

\bibitem[{Pickering and Garrod(2013)}]{pickering2013integrated}
Martin~J Pickering and Simon Garrod. 2013.
\newblock An integrated theory of language production and comprehension.
\newblock \emph{Behavioral and brain sciences}, 36(4):329--347.

\bibitem[{Plepi et~al.(2022)Plepi, Neuendorf, Flek, and
  Welch}]{plepi2022unifying}
Joan Plepi, B{\'e}la Neuendorf, Lucie Flek, and Charles Welch. 2022.
\newblock \href {https://doi.org/10.18653/v1/2022.emnlp-main.500} {Unifying
  data perspectivism and personalization: An application to social norms}.
\newblock In \emph{Proceedings of the 2022 Conference on Empirical Methods in
  Natural Language Processing}, pages 7391--7402, Abu Dhabi, United Arab
  Emirates. Association for Computational Linguistics.

\bibitem[{Poddar et~al.(2023)Poddar, Sinha, Naaman, and
  Jakesch}]{poddar2023topicchange}
Ritika Poddar, Rashmi Sinha, Mor Naaman, and Maurice Jakesch. 2023.
\newblock \href {https://doi.org/10.1145/3544549.3585893} {Ai writing
  assistants influence topic choice in self-presentation}.
\newblock In \emph{Extended Abstracts of the 2023 CHI Conference on Human
  Factors in Computing Systems}, CHI EA '23, New York, NY, USA. Association for
  Computing Machinery.

\bibitem[{Pu and Demberg(2023)}]{pu2023gptvhuman}
Dongqi Pu and Vera Demberg. 2023.
\newblock \href {https://doi.org/10.18653/v1/2023.acl-srw.1} {{C}hat{GPT} vs
  human-authored text: Insights into controllable text summarization and
  sentence style transfer}.
\newblock In \emph{Proceedings of the 61st Annual Meeting of the Association
  for Computational Linguistics (Volume 4: Student Research Workshop)}, pages
  1--18, Toronto, Canada. Association for Computational Linguistics.

\bibitem[{Rubin et~al.(2022)Rubin, Herzig, and Berant}]{rubin2022promptret}
Ohad Rubin, Jonathan Herzig, and Jonathan Berant. 2022.
\newblock \href {https://doi.org/10.18653/v1/2022.naacl-main.191} {Learning to
  retrieve prompts for in-context learning}.
\newblock In \emph{Proceedings of the 2022 Conference of the North American
  Chapter of the Association for Computational Linguistics: Human Language
  Technologies}, pages 2655--2671, Seattle, United States. Association for
  Computational Linguistics.

\bibitem[{Sachan et~al.(2022)Sachan, Lewis, Joshi, Aghajanyan, Yih, Pineau, and
  Zettlemoyer}]{sachan2022upr}
Devendra Sachan, Mike Lewis, Mandar Joshi, Armen Aghajanyan, Wen-tau Yih,
  Joelle Pineau, and Luke Zettlemoyer. 2022.
\newblock \href {https://aclanthology.org/2022.emnlp-main.249} {Improving
  passage retrieval with zero-shot question generation}.
\newblock In \emph{Proceedings of the 2022 Conference on Empirical Methods in
  Natural Language Processing}, pages 3781--3797, Abu Dhabi, United Arab
  Emirates. Association for Computational Linguistics.

\bibitem[{Salemi et~al.(2024)Salemi, Kallumadi, and Zamani}]{salemi2024ropg}
Alireza Salemi, Surya Kallumadi, and Hamed Zamani. 2024.
\newblock \href {https://doi.org/10.1145/3626772.3657783} {Optimization methods
  for personalizing large language models through retrieval augmentation}.
\newblock In \emph{Proceedings of the 47th International ACM SIGIR Conference
  on Research and Development in Information Retrieval}, SIGIR '24, page
  752–762, New York, NY, USA. Association for Computing Machinery.

\bibitem[{Salemi et~al.(2023)Salemi, Mysore, Bendersky, and
  Zamani}]{salemi2023lamp}
Alireza Salemi, Sheshera Mysore, Michael Bendersky, and Hamed Zamani. 2023.
\newblock \href {http://arxiv.org/abs/2304.11406} {Lamp: When large language
  models meet personalization}.

\bibitem[{Shen et~al.(2023)Shen, August, Siangliulue, Lo, Bragg, Hammerbacher,
  Downey, Chang, and Sontag}]{shen2023summarization}
Zejiang Shen, Tal August, Pao Siangliulue, Kyle Lo, Jonathan Bragg, Jeff
  Hammerbacher, Doug Downey, Joseph~Chee Chang, and David Sontag. 2023.
\newblock \href {http://arxiv.org/abs/2304.02623} {Beyond summarization:
  Designing ai support for real-world expository writing tasks}.
\newblock \emph{Second Workshop on Intelligent and Interactive Writing
  Assistants, CHI 2023}.

\bibitem[{Shi et~al.(2023)Shi, Han, Lewis, Tsvetkov, Zettlemoyer, and tau
  Yih}]{shi2023trusting}
Weijia Shi, Xiaochuang Han, Mike Lewis, Yulia Tsvetkov, Luke Zettlemoyer, and
  Scott~Wen tau Yih. 2023.
\newblock \href {http://arxiv.org/abs/2305.14739} {Trusting your evidence:
  Hallucinate less with context-aware decoding}.
\newblock \emph{arXiv preprint arXiv:2305.14739}.

\bibitem[{Shridhar et~al.(2023)Shridhar, Jhamtani, Fang, Durme, Eisner, and
  Xia}]{shridhar2023screws}
Kumar Shridhar, Harsh Jhamtani, Hao Fang, Benjamin~Van Durme, Jason Eisner, and
  Patrick Xia. 2023.
\newblock \href {http://arxiv.org/abs/2309.13075} {Screws: A modular framework
  for reasoning with revisions}.
\newblock \emph{arXiv preprint arXiv:2309.13075}.

\bibitem[{Song et~al.(2020)Song, Tan, Qin, Lu, and Liu}]{song2020mpnet}
Kaitao Song, Xu~Tan, Tao Qin, Jianfeng Lu, and Tie-Yan Liu. 2020.
\newblock \href
  {https://proceedings.neurips.cc/paper_files/paper/2020/file/c3a690be93aa602ee2dc0ccab5b7b67e-Paper.pdf}
  {Mpnet: Masked and permuted pre-training for language understanding}.
\newblock In \emph{Advances in Neural Information Processing Systems},
  volume~33.

\bibitem[{Sun et~al.(2021)Sun, Zhao, Manjunatha, Jain, Morariu, Dernoncourt,
  Srinivasan, and Iyyer}]{sun2021iga}
Simeng Sun, Wenlong Zhao, Varun Manjunatha, Rajiv Jain, Vlad Morariu, Franck
  Dernoncourt, Balaji~Vasan Srinivasan, and Mohit Iyyer. 2021.
\newblock \href {https://doi.org/10.18653/v1/2021.emnlp-main.483} {{IGA}: An
  intent-guided authoring assistant}.
\newblock In \emph{Proceedings of the 2021 Conference on Empirical Methods in
  Natural Language Processing}, pages 5972--5985, Online and Punta Cana,
  Dominican Republic. Association for Computational Linguistics.

\bibitem[{Thakur et~al.(2021)Thakur, Reimers, R{\"u}ckl{\'e}, Srivastava, and
  Gurevych}]{thakur2021beir}
Nandan Thakur, Nils Reimers, Andreas R{\"u}ckl{\'e}, Abhishek Srivastava, and
  Iryna Gurevych. 2021.
\newblock \href {https://openreview.net/forum?id=wCu6T5xFjeJ} {{BEIR}: A
  heterogeneous benchmark for zero-shot evaluation of information retrieval
  models}.
\newblock In \emph{Thirty-fifth Conference on Neural Information Processing
  Systems Datasets and Benchmarks Track (Round 2)}.

\bibitem[{Trajanovski et~al.(2021)Trajanovski, Atalla, Kim, Agarwal, Shokouhi,
  and Quirk}]{trajanovski2021microsoftcontext}
Stojan Trajanovski, Chad Atalla, Kunho Kim, Vipul Agarwal, Milad Shokouhi, and
  Chris Quirk. 2021.
\newblock \href {https://doi.org/10.18653/v1/2021.naacl-industry.1} {When does
  text prediction benefit from additional context? an exploration of contextual
  signals for chat and email messages}.
\newblock In \emph{Proceedings of the 2021 Conference of the North American
  Chapter of the Association for Computational Linguistics: Human Language
  Technologies: Industry Papers}, pages 1--9, Online. Association for
  Computational Linguistics.

\bibitem[{Wang et~al.(2023{\natexlab{a}})Wang, Yang, Zhu, Yang, Cohen, Li, and
  Tian}]{wang2023personalizedeval}
Danqing Wang, Kevin Yang, Hanlin Zhu, Xiaomeng Yang, Andrew Cohen, Lei Li, and
  Yuandong Tian. 2023{\natexlab{a}}.
\newblock \href {http://arxiv.org/abs/2310.03304} {Learning personalized story
  evaluation}.
\newblock \emph{arXiv preprint arXiv:2310.03304}.

\bibitem[{Wang et~al.(2023{\natexlab{b}})Wang, Yang, and
  Wei}]{wang2023learning}
Liang Wang, Nan Yang, and Furu Wei. 2023{\natexlab{b}}.
\newblock \href {http://arxiv.org/abs/2307.07164} {Learning to retrieve
  in-context examples for large language models}.
\newblock \emph{arXiv preprint arXiv:2307.07164}.

\bibitem[{Wang et~al.(2023{\natexlab{c}})Wang, Chilton, and
  Nickerson}]{wang2023writingjourn}
Sitong Wang, Lydia~B Chilton, and Jeffrey~V Nickerson. 2023{\natexlab{c}}.
\newblock \href
  {https://cdn.glitch.global/d058c114-3406-43be-8a3c-d3afff35eda2/paper25_2023.pdf}
  {Writing with generative ai: Multi-modal and multi-dimensional tools for
  journalists}.
\newblock \emph{Second Workshop on Intelligent and Interactive Writing
  Assistants, CHI 2023}.

\bibitem[{Wang et~al.(2023{\natexlab{d}})Wang, Jiang, Zhang, Li, Liang, Mei,
  and Bendersky}]{wang2023automated}
Yaqing Wang, Jiepu Jiang, Mingyang Zhang, Cheng Li, Yi~Liang, Qiaozhu Mei, and
  Michael Bendersky. 2023{\natexlab{d}}.
\newblock \href {http://arxiv.org/abs/2310.11593} {Automated evaluation of
  personalized text generation using large language models}.
\newblock \emph{arXiv preprint arXiv:2310.11593}.

\bibitem[{Wu et~al.(2022)Wu, Terry, and Cai}]{tongshuang2022aichains}
Tongshuang Wu, Michael Terry, and Carrie~Jun Cai. 2022.
\newblock \href {https://doi.org/10.1145/3491102.3517582} {Ai chains:
  Transparent and controllable human-ai interaction by chaining large language
  model prompts}.
\newblock In \emph{Proceedings of the 2022 CHI Conference on Human Factors in
  Computing Systems}, CHI '22, New York, NY, USA. Association for Computing
  Machinery.

\bibitem[{Yadav et~al.(2022)Yadav, Monath, Angell, Zaheer, and
  McCallum}]{yadav2022anncur}
Nishant Yadav, Nicholas Monath, Rico Angell, Manzil Zaheer, and Andrew
  McCallum. 2022.
\newblock \href {https://doi.org/10.18653/v1/2022.emnlp-main.140} {Efficient
  nearest neighbor search for cross-encoder models using matrix factorization}.
\newblock In \emph{Proceedings of the 2022 Conference on Empirical Methods in
  Natural Language Processing}, pages 2171--2194, Abu Dhabi, United Arab
  Emirates. Association for Computational Linguistics.

\bibitem[{Yan et~al.(2022)Yan, Qin, Wang, Bendersky, and
  Najork}]{le2022scalecalib}
Le~Yan, Zhen Qin, Xuanhui Wang, Michael Bendersky, and Marc Najork. 2022.
\newblock \href {https://doi.org/10.1145/3534678.3539072} {Scale calibration of
  deep ranking models}.
\newblock In \emph{Proceedings of the 28th ACM SIGKDD Conference on Knowledge
  Discovery and Data Mining}, KDD '22, page 4300–4309, New York, NY, USA.
  Association for Computing Machinery.

\bibitem[{Yoran et~al.(2023)Yoran, Wolfson, Ram, and Berant}]{yoran2023making}
Ori Yoran, Tomer Wolfson, Ori Ram, and Jonathan Berant. 2023.
\newblock \href {http://arxiv.org/abs/2310.01558} {Making retrieval-augmented
  language models robust to irrelevant context}.

\bibitem[{Zamani and Bendersky(2023)}]{zamani2023multivar}
Hamed Zamani and Michael Bendersky. 2023.
\newblock \href {https://doi.org/10.1145/3539618.3591740} {Multivariate
  representation learning for information retrieval}.
\newblock In \emph{Proceedings of the 46th International ACM SIGIR Conference
  on Research and Development in Information Retrieval}, SIGIR '23, page
  163–173, New York, NY, USA. Association for Computing Machinery.

\bibitem[{Zhang et~al.(2023{\natexlab{a}})Zhang, Song, Li, Zhou, and
  Song}]{zhang2023ctrlgen}
Hanqing Zhang, Haolin Song, Shaoyu Li, Ming Zhou, and Dawei Song.
  2023{\natexlab{a}}.
\newblock \href {https://doi.org/10.1145/3617680} {A survey of controllable
  text generation using transformer-based pre-trained language models}.
\newblock \emph{ACM Comput. Surv.}, 56(3).

\bibitem[{Zhang et~al.(2023{\natexlab{b}})Zhang, Krishna, Awadallah, and
  Wang}]{zhang2023ecoassistant}
Jieyu Zhang, Ranjay Krishna, Ahmed~H. Awadallah, and Chi Wang.
  2023{\natexlab{b}}.
\newblock \href {http://arxiv.org/abs/2310.03046} {Ecoassistant: Using llm
  assistant more affordably and accurately}.
\newblock \emph{arXiv preprint arXiv:2310.03046}.

\bibitem[{Zhang et~al.(2018)Zhang, Dinan, Urbanek, Szlam, Kiela, and
  Weston}]{zhang2018personalizeddial}
Saizheng Zhang, Emily Dinan, Jack Urbanek, Arthur Szlam, Douwe Kiela, and Jason
  Weston. 2018.
\newblock \href {https://doi.org/10.18653/v1/P18-1205} {Personalizing dialogue
  agents: {I} have a dog, do you have pets too?}
\newblock In \emph{Proceedings of the 56th Annual Meeting of the Association
  for Computational Linguistics (Volume 1: Long Papers)}, pages 2204--2213,
  Melbourne, Australia. Association for Computational Linguistics.

\bibitem[{Zhang* et~al.(2020)Zhang*, Kishore*, Wu*, Weinberger, and
  Artzi}]{Zhang2020BERTScore}
Tianyi Zhang*, Varsha Kishore*, Felix Wu*, Kilian~Q. Weinberger, and Yoav
  Artzi. 2020.
\newblock \href {https://openreview.net/forum?id=SkeHuCVFDr} {Bertscore:
  Evaluating text generation with bert}.
\newblock In \emph{International Conference on Learning Representations}.

\bibitem[{Zhang et~al.(2022)Zhang, Feng, and Tan}]{zhang2022active}
Yiming Zhang, Shi Feng, and Chenhao Tan. 2022.
\newblock \href {https://doi.org/10.18653/v1/2022.emnlp-main.622} {Active
  example selection for in-context learning}.
\newblock In \emph{Proceedings of the 2022 Conference on Empirical Methods in
  Natural Language Processing}, pages 9134--9148, Abu Dhabi, United Arab
  Emirates. Association for Computational Linguistics.

\end{thebibliography}

\appendix

\section{Model Details}
\label{supp-implementation}
\begin{lstfloat}[t]
\begin{lstlisting}[breakautoindent=false, breaklines=true, breakindent=0pt]
For a +\textcolor{BrickRed}{POST}+ from the subreddit Am I The Asshole write a +\textcolor{BrickRed}{COMMENT}+ explaining if the author of a post is an asshole or not the asshole as a +\textcolor{BrickRed}{COMMENTER}+. 
Use the following instructions for your response:
1. Read the below example comments by the +\textcolor{BrickRed}{COMMENTER}+.
2. Write the comment as the +\textcolor{BrickRed}{COMMENTER}+ mimicing the length, style, reasoning, and stances of their comments.
Here are some example comments by the +\textcolor{BrickRed}{COMMENTER}+: +\textcolor{Fuchsia}{\textbf{\{\{historical\_examples\}\}}}+
+\textcolor{BrickRed}{POST}+: +\textcolor{Fuchsia}{\textbf{\{\{target\_request\}\}}}+
Write the +\textcolor{BrickRed}{COMMENT}+ mimicing the length, style, reasoning, and stances of the +\textcolor{BrickRed}{COMMENTERS}+ comments.
\end{lstlisting}
\caption{$\llmG$ prompt used to generate a target text given historical examples retrieved by $\retriever$ and a target request for \aita.}
\label{fig-aita-copy-prompt}
\end{lstfloat}
\begin{lstfloat}[t]
\begin{lstlisting}[breakautoindent=false, breaklines=true, breakindent=0pt]
Given a +\textcolor{BrickRed}{REQUEST}+ from a +\textcolor{BrickRed}{USER}+ to author a +\textcolor{BrickRed}{POST}+, write a +\textcolor{BrickRed}{POST}+ for an enterprise social media site mimicking the user to satisfy the +\textcolor{BrickRed}{REQUEST}+.
Use the following instructions for your response:
1. You should maintain consistency in tone and style with the +\textcolor{BrickRed}{USER}+'s historical posts.
2. You should imitate the language style of the +\textcolor{BrickRed}{USER}+'s historical posts.
3. You should employ similar rhetorical methods as the +\textcolor{BrickRed}{USER}+'s historical posts.
Here are some historical posts by the +\textcolor{BrickRed}{USER}+: +\textcolor{Fuchsia}{\textbf{\{\{historical\_examples\}\}}}+
+\textcolor{BrickRed}{REQUEST}+: +\textcolor{Fuchsia}{\textbf{\{\{target\_request\}\}}}+
Write the +\textcolor{BrickRed}{POST}+ to satisfy the +\textcolor{BrickRed}{REQUEST}+ mimicing the tone, style, and rhetorical methods of the +\textcolor{BrickRed}{USER}+'s historical posts.
\end{lstlisting}
\caption{$\llmG$ prompt used to generate a target text given historical examples retrieved by $\retriever$ and a target request for \worksm.}
\label{fig-worksm-copy-prompt}
\end{lstfloat}
\textbf{Retriever}
We instantiate $\retriever$ with the pre-trained \textsc{MPnet}, which is relatively lightweight at 110M parameters~\citep{song2020mpnet}. 
We obtain an output score from $\retriever$ as $\mathbf{w}^T\texttt{tanh}\left(\mathbf{W}^T\textsc{Enc}([{q}_u, {d}_u])\right)$, where \textsc{Enc} represents the \textsc{cls} token from the final layer of the encoder, and ${q}_u$ and ${d}_u$ are the text of the input request and historical document. 
The encoder parameters, $\mathbf{w}$, and $\mathbf{W}$ are trained.

\textbf{Text generation models}
For $\llmG$ we consider two performant LLMs offered via API by Azure OpenAI, \texttt{davinci-003} and \texttt{gpt-3.5-turbo}.
For $\slm$ we consider a smaller but still effective encoder-decoder language model, \textsc{FlanT5-XL}, with 3 billion parameters~\citep{chung2022scaling}. 
The latter model is open-sourced, allowing us to access its token likelihoods directly, a requirement of Eq.~\ref{eq-diff-likelihood}. 
We obtain target text likelihoods by taking the average of log-probabilities of individual token likelihoods from \textsc{FlanT5-XL}.

\textbf{LLM prompts} We use Prompts \ref{fig-worksm-copy-prompt} and \ref{fig-aita-copy-prompt} for LLM inference. The same prompts are used with \texttt{davinci-003} and \texttt{gpt-35-turbo}. For constructing training data in Eq \ref{eq-diff-likelihood} with a \textsc{FlanT5-XL}, $\slm$ we use Prompts \ref{fig-worksm-faux-global-prompt}, \ref{fig-worksm-faux-personal-prompt}, \ref{fig-aita-faux-global-prompt}, \ref{fig-aita-faux-personal-prompt}. Note that computing $p_{\textrm{aux}}(\target | \request)$ uses a set of randomly chosen few shot examples from the training set fixed across requests rather than the request alone.

\textbf{\proposedmethod implementation}
In constructing training data for $\retriever$ we use $|\historicalposts^t|=8$, i.e we treat the 8 most recent texts per user as their target texts. To train $\retriever$, we consider the top two canadidate documents per Eq.~\eqref{eq-diff-likelihood} as positive examples per request and use three negatives per positive, i.e.,  $P=2$ and $N=3$. In our LLM prompts, we use $k=3$ retrieved examples for \worksm and $k=4$ for \aita, tuned  on a dev set, and set generation temperature to zero. 

We also use temperatures for target scores input to softmax functions in Eq.~\eqref{eq-kl-loss}, $\mathbf{y}_u'/\tau$ with $\tau=5$. Finally, we set $y_0=110$ for \worksm and $y_0=5$ for \aita, which are the median values of Eq.~\eqref{eq-diff-likelihood} for each respective dataset on the training data. We tuned $y_0$ on a dev set constructed similar to our training set to 25 and 75 percentile values of Eq.~\eqref{eq-diff-likelihood}. Our retrievers were trained on Nvidia V100 GPUs with 16GB memory or Nvidia RTX A6000 GPUs with 48GB memory. Experiments for training retrievers required about 300 hours in total.
\begin{lstfloat}[t]
\begin{lstlisting}[breakautoindent=false, breaklines=true, breakindent=0pt]
Here are some example posts on the Am I The Asshole subreddit: +\textcolor{Fuchsia}{\textbf{\{\{random\_fewshot\_examples\}\}}}+. Target post: +\textcolor{Fuchsia}{\textbf{\{\{target\_post\}\}}}+. Write a users comment for this post:
\end{lstlisting}
\caption{$\slm$ prompt used to compute $p_{\textrm{aux}}(\target | \request)$ in Eq \eqref{eq-diff-likelihood} for \aita.}
\label{fig-aita-faux-global-prompt}
\end{lstfloat}
\begin{lstfloat}[t]
\begin{lstlisting}[breakautoindent=false, breaklines=true, breakindent=0pt]
Here is an comment on a post by a user on the Am I the Asshole subreddit: +\textcolor{Fuchsia}{\textbf{\{\{candidate\_comment\}\}}}+. Target post: +\textcolor{Fuchsia}{\textbf{\{\{target\_post\}\}}}+. Write a users comment for this post:
\end{lstlisting}
\caption{$\slm$ prompt used to compute $p_{\textrm{aux}}(\target | \doc, \request)$ in Eq \eqref{eq-diff-likelihood} for \aita.}
\label{fig-aita-faux-personal-prompt}
\end{lstfloat}
\begin{lstfloat}[t]
\begin{lstlisting}[breakautoindent=false, breaklines=true, breakindent=0pt]
Here is are some posts by a user on an enterprise social network: +\textcolor{Fuchsia}{\textbf{\{\{random\_fewshot\_examples\}\}}}+
Here is an outline for a target post by the user: +\textcolor{Fuchsia}{\textbf{\{\{target\_request\}\}}}+. Write the target post:
\end{lstlisting}
\caption{$\slm$ prompt used to compute $p_{\textrm{aux}}(\target | \request)$ in Eq \eqref{eq-diff-likelihood} for \worksm.}
\label{fig-worksm-faux-global-prompt}
\end{lstfloat}
\begin{lstfloat}[t]
\begin{lstlisting}[breakautoindent=false, breaklines=true, breakindent=0pt]
Here is an example post by a user on an enterprise social network: +\textcolor{Fuchsia}{\textbf{\{\{candidate\_document\}\}}}+. Here is an outline for a target post by the user: +\textcolor{Fuchsia}{\textbf{\{\{target\_request\}\}}}+. Write the target post:
\end{lstlisting}
\caption{$\slm$ prompt used to compute $p_{\textrm{aux}}(\target | \doc, \request)$ in Eq \eqref{eq-diff-likelihood} for \worksm.}
\label{fig-worksm-faux-personal-prompt}
\end{lstfloat}

\section{Experimental Details}
\label{supp-datasets}
Here we present various details of datasets, baselines, and manual evaluation.



\subsection{Evaluation Requests in \worksm}
For evaluation in \worksm two authors not involved in model development manually authored requests for each of the 163 target posts in our evaluation set. Guidelines presented to annotators for the requests are presented in Guideline \ref{fig-req-guidelines}. The requests are intended to be brief and include the salient information contained in the post. Note that annotators external to the authors weren't recruited for authoring requests due to the private and highly regulated nature of \worksm.

\subsection{Training Requests in \worksm} 
Section \ref{exp:data} notes that our training set for \worksm was constructed from synthetic requests generated by GPT4. The prompt for this is presented in Prompt \ref{fig-worksm-reqgen}. We follow an incremental approach to construct the synthetic requests: first extracting the salient aspects of the post, followed by concatenation of these aspects to result in the request. The salient aspects span: an overview of the post, proper nouns mentioned in the post, contact information, links to webpages, and any specialized knowledge or anecdotes in the post. Given the success of chain-of-thought prompting, we generate an explanation followed by salient aspects of the post -- the explanations are not used elsewhere. Enterprise contracts ensure the privacy of user data shared over the API.

\subsection{Baselines}
\label{supp-baselines}
We consider the following non-personalized baselines:
\ul{\textsc{zShot-NP}}: This represents a non-personalized approach prompting only with the request. \ul{\textsc{kShot-NP}}: A zero-shot non-personalized approach using a fixed randomly selected set of $k$ documents for all requests. For \aita, the examples are balanced across labels. 

We consider the following retrieval-augmented personalized baselines, selecting from a user's historical documents $\historicalposts$: \ul{Random}: Random selection of $k$ documents from $\historicalposts$. \ul{BM25}: Represents a classic performant retrieval model based on query-document term overlap. \ul{\textsc{MPNet-1B}}: This a strong \textsc{MPNet} bi-encoder trained on 1 billion text pairs from numerous domains.\footnote{HF model: \texttt{sentence-transformers/all-mpnet-base-v2}} Documents are ranked for a request using cosine similarity between embeddings. \ul{QL-FT5}: An approach which ranks documents based on $p(\request | \doc)$ with a pretrained \textsc{FlanT5-Base} with 250M parameters ~\citep{sachan2022upr}. This may be seen as an unsupervised crossencoder. \ul{RelevanceCE}: A supervised crossencoder with the same architecture as $\retriever$ in \proposedmethod but differing in training. This is trained on pairs of $(\request, \doc)$ in $\historicalposts$ treated as positive training pairs with a crossentropy loss, with negatives selected as a random historical document from the same user not but corresponding to $\request$. Note that this corresponds to a crossencoder optimized for request-document relevance, i.e. $p(\text{relevance}=1|\request, \doc)$, rather than personalized target text generation. 

\subsection{Judge LLM prompts}
\label{sec-exp-judgellmprompt}
In Prompt \ref{fig-aita-judgellm-prompt} and \ref{fig-worksm-judgellm-prompt} we present prompts for GPT-4o as a judge LLM discussed in \S\ref{sec-main-results}.
\begin{lstfloat}[t]
\begin{lstlisting}[breakautoindent=false, breaklines=true, breakindent=0pt]
You an an experienced linguist who helps people compare social media texts.
Given a +\textcolor{BrickRed}{REFERENCE POST}+ and two 
+\textcolor{BrickRed}{TARGET POSTS}+ judge which of the TARGET POSTs is significantly more likely to be 
written by the same author as the +\textcolor{BrickRed}{REFERENCE POST}+. 
For your response use the following instructions:
1. Make your judgement based on stylistic patterns, ordering of information, and tone used.
2. Output +\textcolor{BrickRed}{POST ONE}+ if it is significantly more likely to be written by the same author as the +\textcolor{BrickRed}{REFERENCE POST}+.
3. Output +\textcolor{BrickRed}{POST TWO}+ if it is significantly more likely to be written by the same author as the +\textcolor{BrickRed}{REFERENCE POST}+.
4. Output BOTH if either post could have been written by the same author or neither could have been written by the same author.
Here are the POSTS:
+\textcolor{BrickRed}{REFERENCE POST}+: +\textcolor{Fuchsia}{\textbf{\{\{reference\_post\}\}}}+
+\textcolor{BrickRed}{POST ONE}+: +\textcolor{Fuchsia}{\textbf{\{\{post\_one\}\}}}+
+\textcolor{BrickRed}{POST TWO}+: +\textcolor{Fuchsia}{\textbf{\{\{post\_two\}\}}}+
Output a justification for your judgement, then output POST ONE, POST TWO, or BOTH to indicate your final decision.
\end{lstlisting}
\caption{Judge LLM prompt used to select a generated post more likely to align with a reference post authored by a user for \worksm.}
\label{fig-worksm-judgellm-prompt}
\end{lstfloat}
\begin{lstfloat}[t]
\begin{lstlisting}[breakautoindent=false, breaklines=true, breakindent=0pt]
You an an experienced linguist who helps people compare social media texts.
Given a +\textcolor{BrickRed}{REFERENCE POST}+ and two 
+\textcolor{BrickRed}{TARGET POSTS}+ judge which of the TARGET POSTs is significantly more likely to be 
written by the same author as the +\textcolor{BrickRed}{REFERENCE POST}+. 
For your response use the following instructions:
1. Make your judgement based on similarity of stylistic patterns, arguments, stances, and word choices.
2. Output +\textcolor{BrickRed}{POST ONE}+ if it is significantly more likely to be written by the same author as the +\textcolor{BrickRed}{REFERENCE POST}+.
3. Output +\textcolor{BrickRed}{POST TWO}+ if it is significantly more likely to be written by the same author as the +\textcolor{BrickRed}{REFERENCE POST}+.
4. Output BOTH if either post could have been written by the same author or neither could have been written by the same author.
Here are the POSTS:
+\textcolor{BrickRed}{REFERENCE POST}+: +\textcolor{Fuchsia}{\textbf{\{\{reference\_post\}\}}}+
+\textcolor{BrickRed}{POST ONE}+: +\textcolor{Fuchsia}{\textbf{\{\{post\_one\}\}}}+
+\textcolor{BrickRed}{POST TWO}+: +\textcolor{Fuchsia}{\textbf{\{\{post\_two\}\}}}+
Output a justification for your judgement, then output POST ONE, POST TWO, or BOTH to indicate your final decision.
\end{lstlisting}
\caption{Judge LLM prompt used to select a generated comment more likely to align with a reference comment authored by a user for \aita.}
\label{fig-aita-judgellm-prompt}
\end{lstfloat}

\section{Additional Results}
\label{supp-additional-res}
Here we present additional results in addition to those presented in \S\ref{sec-main-results}. We present these here primarily in the interest of space.

\subsection{Qualitative example from \aita}
\begin{figure*}[t]
     \centering
     \includegraphics[width=0.95\textwidth]{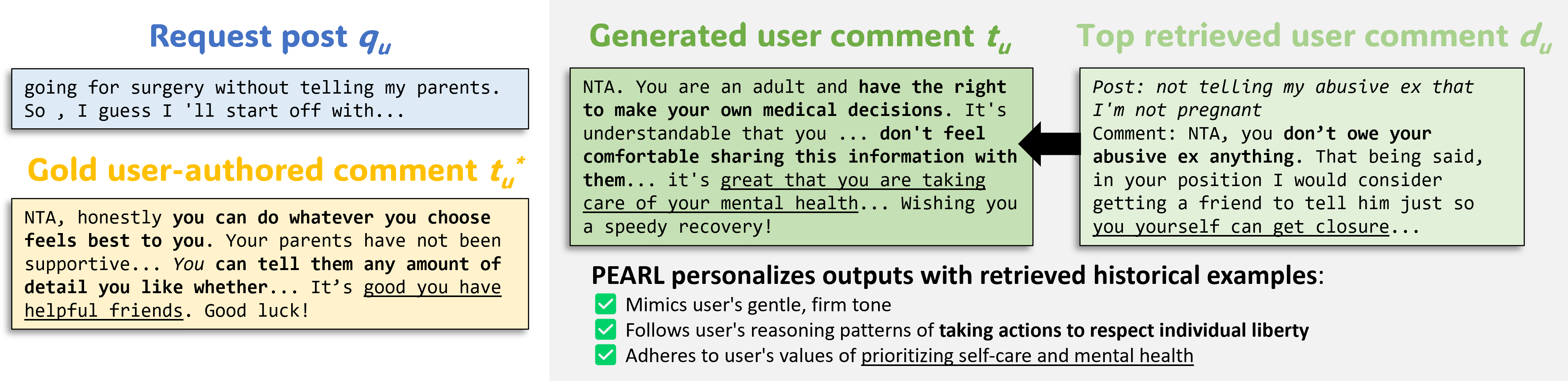}
     \caption{A qualitative example illustrating the effectiveness of  PEARL on \aita: Given a request post $\request$ describing an ambiguous interpersonal situation regarding sharing medical information, PEARL retrieves a historical user comment $\doc$ that demonstrates the user's characteristic tone and values, and generates a comment $t_u$ highly similar to the ground-truth user comment $t_u^*$. 
     We \textbf{bold} qualitatively similar phrases about individual liberties and \emph{italicize} phrases about self-care and mental health. All texts are abbreviated for space, and provided in full in Appendix~\ref{supp-additional-res}.}
     \label{fig-pearl-example}
 \end{figure*}
Here we provide the full texts of the qualitative \aita examples from Figure~\ref{fig-pearl-example}, alongside the top-retrieved comment from BM25 to show the qualitative difference in retriever effectiveness. 

\textbf{Original post $\request$}:
going for surgery without telling my parents. So , I guess I 'll start off with background information .I am 21 , and
trans
masculine , or female to male if that term works for you

\textbf{Gold user-authored comment $t_u^{*}$}:
``NTA, honestly you can do whatever you choose feels best to you. Your parents have not been supportive of your transition so they have elected not to be a part of the process. You can tell them any amount of detail you like whether before, after, or at different times. It’s good you have helpful friends. Good luck!''

\textbf{Top \proposedmethod-retrieved comment $\doc$}:
``NTA, you don’t owe your abusive ex anything. That being said, in your position I would consider getting a friend to tell him just so you yourself can get closure and not be wondering when he’s going to harass you about it.''

\textbf{Top BM25-retrieved comment $\doc$}: 
``YTA, just another lame guy who is only interested in a girl if he has a chance to sleep with her''

\textbf{\proposedmethod LLM output}: 
``NTA. You are an adult and have the right to make your own medical decisions. It's understandable that you don't have a great relationship with your parents and don't feel comfortable sharing this information with them. It's also important to note that being
transgender is not self-harm and it's great that you are taking care of your mental health by seeing a counselor and nurse regularly. Wishing you a speedy recovery!''

\subsection{Ablating Training Strategies}
\label{sec-ablations}
In Table \ref{tab-both-gpt35t-ablations} and \ref{tab-both-davinci-ablations}, we compare common alternative training of $\retriever$ while keeping inference steps identical for \texttt{gpt-35-turbo} and \texttt{davinci-003} respectively. These serve to ablate our specific contributions: selection of training requests which benefit from personalization and our scale calibrating objective.

\textbf{No calibrated objective}
Removing the scale calibration and using a standard KL divergence (-- calibrated \texttt{sm}) degrades performance for \aita and sees comparable performance in \worksm with \texttt{gpt-35-turbo} in Table \ref{tab-both-gpt35t-ablations}. With \texttt{davinci-003} we see scale calibration consistently improves performance (Table \ref{tab-both-davinci-ablations}).
This indicates the importance of calibration for estimating the benefit of a historical document to a request consistently across datasets and LLMs. Appendix \ref{sec-calibration-eval} shows scale calibration also consistently improves the correlation of retriever scores with task performance.
\begin{table}[]
\centering
\scalebox{0.71}{
\begin{tabular}{rcccc}
\toprule
Dataset $\rightarrow$ & {\aita} & \multicolumn{3}{c}{\worksm}\\
\cmidrule(lr){2-2}\cmidrule(lr){3-5}
LLM $\rightarrow$ & \texttt{gpt-35-turbo} & \multicolumn{3}{c}{\texttt{gpt-35-turbo}}\\
\cmidrule(lr){2-2} \cmidrule(lr){3-5}
Method $\downarrow$ & Macro F1(\%) & \small{\textsc{BS-F1}} & \small{\textsc{R1}} & \small{\textsc{R2}}\\
\toprule
\proposedmethod  & \textbf{\underline{65.34}} & \underline{36.49} & \underline{0.5082}	& \textbf{\underline{0.2676}}\\
\midrule
no calibrated \texttt{sm} & {63.01} & \textbf{\underline{36.69}} & \textbf{\underline{0.5095}} & \underline{0.2654}\\
distill $p_{\textrm{aux}} (\target | \doc, \request)$ & 62.32 & 35.64 & 0.5057 & 0.2652\\
\bottomrule
\end{tabular}}
\caption{\proposedmethod compared to alternate training of $\retriever$ for \texttt{gpt-35-turbo}.}
\label{tab-both-gpt35t-ablations}
\end{table}
\begin{table}[]
\centering
\scalebox{0.71}{
\begin{tabular}{rcccc}
\toprule
Dataset $\rightarrow$ & {\aita} & \multicolumn{3}{c}{\worksm}\\
\cmidrule(lr){2-2}\cmidrule(lr){3-5}
LLM $\rightarrow$ & \texttt{davinci-003} & \multicolumn{3}{c}{\texttt{davinci-003}}\\
\cmidrule(lr){2-2} \cmidrule(lr){3-5}
Method $\downarrow$ & Macro F1(\%) & \small{\textsc{BS-F1}} & \small{\textsc{R1}} & \small{\textsc{R2}}\\
\toprule
\proposedmethod & \textbf{\underline{61.21}} & \textbf{\underline{39.60}} & \textbf{\underline{0.5419}} & \textbf{0.3094}\\
\midrule
no calibrated \texttt{sm} & 57.27 & {38.88} & {0.5350} & {0.3033}\\
distill $p_{\textrm{aux}} (\target | \doc, \request)$ & 55.52 & 39.34 & 0.5359	& 0.3059\\
\bottomrule
\end{tabular}}
\caption{\proposedmethod compared to alternate training of $\retriever$ for \texttt{davinci-003}.}
\label{tab-both-davinci-ablations}
\end{table}

\textbf{Distill $p_{\textrm{aux}} (\target | \doc, \request)$ to $\retriever$.} The proposed $\retriever$ is trained on documents which benefit personalization \emph{and} requests which benefit from personalization.
Here, we compare to an approach that only selects documents that benefit personalization by maximizing $p_\textrm{aux}(t_u|q_u,d_u)$. This assumes that \textit{all} training requests benefit from personalization. We train $\retriever$ with a KL-divergence objective. This approach, also, closely resembles prior work example selection in non-personalized tasks \citep{rubin2022promptret} as well as personalized tasks \cite{salemi2024ropg}. We see in Table \ref{tab-both-gpt35t-ablations} and \ref{tab-both-davinci-ablations} (distill $p_{\textrm{aux}} (\target | \doc, \request)$) that this lowers performance markedly, indicating the value of our approach.

\subsection{Calibration Evaluation}
\label{sec-calibration-eval}
Since we aim to train generation calibrated retrievers, we evaluate calibration performance i.e a retrieval models scores to be predictive of downstream generation performance (Table \ref{tab-both-calib}). Here, Pearson $r$ is reported between the top-1 document score for a request and the downstream generation evaluation metric -- {\small R2} for \worksm, and Macro-F1 for \aita. To do this for \aita, we first bin evaluation requests into equal sized bins by top-1 document score, $s_1$, and then measure Pearson $r$ between the bin start and the average evaluation metric per bin. Our metric is in contrast with prior work \citep{dhuliawala2022mrcalibration, le2022scalecalib} that focuses on classification tasks, where model-predicted class probabilities can be used for measuring calibration, missing in our setup.

Among baseline methods, we see sparse and dense retrieval methods, BM25 and \textsc{MPNet-1B} scores to be better calibrated with downstream performance compared to likelihood-based methods like QL-FT5. Next, we see \proposedmethod to be better correlated with downstream performance for \worksm and \aita - indicating the effectiveness of our training strategy. Further, we also report on an ablated model, not using the scale-calibrated objective of Eq \eqref{eq-calibrated-expansion} (-- calibrated \texttt{sm}). We see this approach underperform \proposedmethod, indicating the importance of the scale-calibrated objective for a well-calibrated crossencoder. The poorer calibration of crossencoders also finds support in prior work showing their scores to lie at extremes of the score distribution \citep{menon22defensedual, yadav2022anncur}.
\begin{table}[]
\centering
\scalebox{0.75}{
\begin{tabular}{rrcc}
\toprule
\multicolumn{2}{r}{Method $\downarrow$~/~LLM $\rightarrow$} & \texttt{davinci-003} & \texttt{gpt-35-turbo}\\
             \cmidrule(lr){3-3}\cmidrule(lr){4-4}
& & Pearson $r$ & Pearson $r$ \\
\toprule
\multirow{6}*{\aita}  & BM25 & {0.08} & -0.05 \\
& \textsc{MPNet-1B} & 0.07 & -0.14\\
& \textsc{UPR} & -0.48 & {-0.02}\\
& RelevanceCE & 0.07 & -0.19\\
& \proposedmethod $\retriever$ & \textbf{0.11} & \textbf{0.45}\\
& -- calibrated \texttt{sm} & -0.48 & 0.12\\
\midrule
\multirow{6}*{\worksm} & {BM25} & 0.42 & {0.52}\\
& \textsc{MPNet-1B} & {0.54} & {0.52}\\
& \textsc{UPR} & -0.05 & -0.02\\
& RelevanceCE & 0.56 & 0.49\\
& \proposedmethod $\retriever$ & \textbf{0.64} & \textbf{0.64}\\
& -- calibrated \texttt{sm} & 0.58 & 0.55\\
\bottomrule
\end{tabular}}
\caption{Calibration performance of \proposedmethod evaluated through correlation between score for top-1 document and Macro-F1 for \aita, and {\small R2} for \worksm.}
\label{tab-both-calib}
\end{table}

\subsection{Selective Revision with \proposedmethod ~-- Extended Results}
\label{supp-selective-editing}
In \S\ref{sec-routing-results} we demonstrate how our trained retrieval model can be used for selective revision with \texttt{gpt-35-turbo}. Prompt \ref{fig-aita-revision-prompt}, \ref{fig-worksm-revision-prompt} present the prompts used for revision with both LLMs.
\begin{lstfloat}[t]
\begin{lstlisting}[breakautoindent=false, breaklines=true, breakindent=0pt]
Given a +\textcolor{BrickRed}{POST}+ from the subreddit Am I the Asshole and a +\textcolor{BrickRed}{DRAFT}+ comment from the +\textcolor{BrickRed}{USER}+ responding if the author of the POST is an asshole or not the asshole, edit the +\textcolor{BrickRed}{DRAFT}+ comment.
Use the following instructions for your response:
1. Maintain consistency in tone and style with the USER's historical comments.
2. Edit the +\textcolor{BrickRed}{DRAFT}+ to use more reddit lingo.
3. Remove statements of the +\textcolor{BrickRed}{POST}+ from the +\textcolor{BrickRed}{DRAFT}+.
4. Output a justification for your edits starting with the word JUSTIFICATION.
5. Output the edited +\textcolor{BrickRed}{DRAFT}+ comment starting with the words EDITED DRAFT.
Here are some historical comments by the USER: +\textcolor{Fuchsia}{\textbf{\{\{historical\_examples\}\}}}+
+\textcolor{BrickRed}{REQUEST}+: +\textcolor{Fuchsia}{\textbf{\{\{target\_request\}\}}}+
+\textcolor{BrickRed}{DRAFT}+: +\textcolor{Fuchsia}{\textbf{\{\{target\_draft\}\}}}+
Output a justification for your edits, then output the edited +\textcolor{BrickRed}{DRAFT}+ starting with the words EDITED DRAFT.
\end{lstlisting}
\caption{$\llmG$ prompt used to for selective revision given a Stage 1 draft for \aita.}
\label{fig-aita-revision-prompt}
\end{lstfloat}
\begin{lstfloat}[t]
\begin{lstlisting}[breakautoindent=false, breaklines=true, breakindent=0pt]
Given a +\textcolor{BrickRed}{REQUEST}+ and a +\textcolor{BrickRed}{DRAFT}+ from a +\textcolor{BrickRed}{USER}+ to author a social media +\textcolor{BrickRed}{POST}+, edit the +\textcolor{BrickRed}{DRAFT}+ to satisfy the +\textcolor{BrickRed}{REQUEST}+.
Use the following instructions for your response:
1. Enumerate any missing missing information from the +\textcolor{BrickRed}{REQUEST}+ in the +\textcolor{BrickRed}{DRAFT}+.
2. Enumerate any irrelevant information for the +\textcolor{BrickRed}{REQUEST}+ in the +\textcolor{BrickRed}{DRAFT}+.
3. Then output the edited +\textcolor{BrickRed}{DRAFT}+ starting with the words EDITED DRAFT. 
+\textcolor{BrickRed}{REQUEST}+:+\textcolor{Fuchsia}{\textbf{\{\{target\_request\}\}}}+
+\textcolor{BrickRed}{DRAFT}+: +\textcolor{Fuchsia}{\textbf{\{\{target\_draft\}\}}}+
Output missing or irrelevant information for the +\textcolor{BrickRed}{REQUEST}+, then output the EDITED DRAFT satisfying the +\textcolor{BrickRed}{REQUEST}+.
\end{lstlisting}
\caption{$\llmG$ prompt used to for selective revision given a Stage 1 draft for \worksm.}
\label{fig-worksm-revision-prompt}
\end{lstfloat}

\begin{figure*}
    \centering
    \subfloat[Effectiveness of selective revision for requests of different lengths (in words).\label{fig-selective-rev-perq-lens}]{
        \centering
        \includegraphics[width=0.45\textwidth]{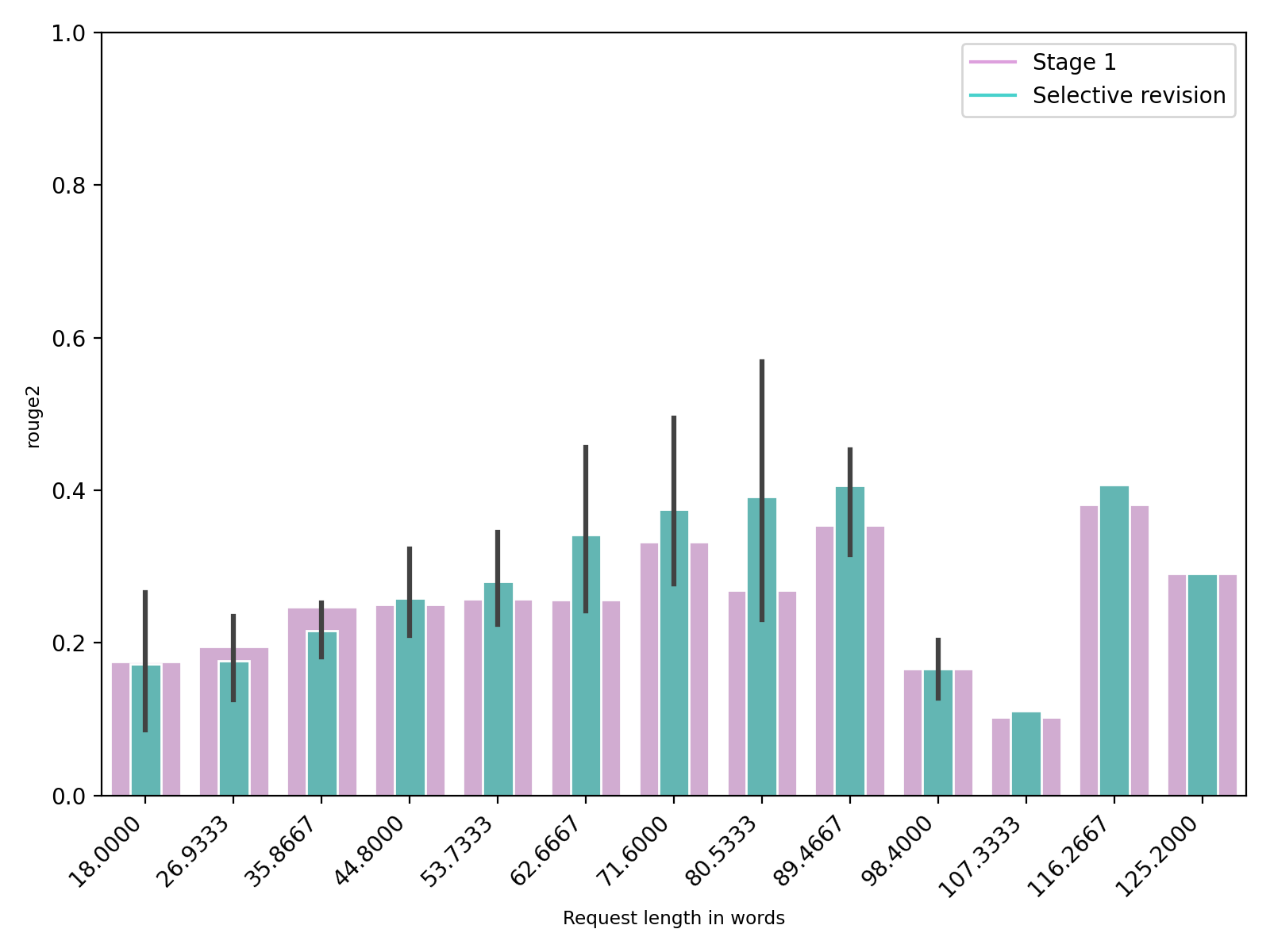}
    }\hfill
    \subfloat[Effectiveness of selective revision for users of different numbers of historical posts.\label{fig-selective-rev-perq-users}]{
        \centering
        \includegraphics[width=0.45\textwidth]{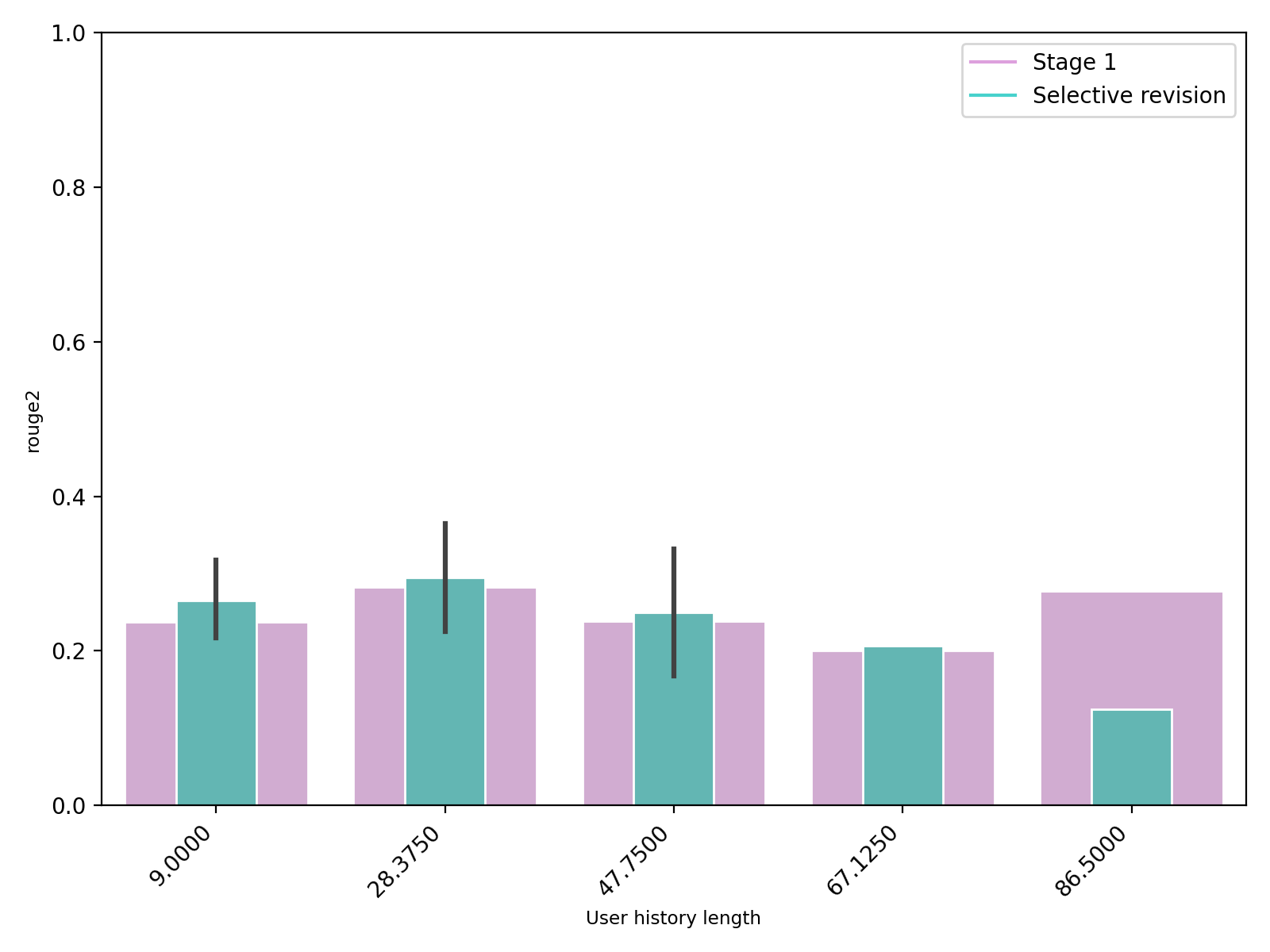}
    }
    \caption{The impact of selective revision (\S\ref{sec-routing-results}) in \proposedmethod on \worksm compared for requests of different length and users with varying number of historical posts.}
    \label{fig-selective-rev-perq}
\end{figure*}

In Figure \ref{fig-selective-rev-perq}, we examine the impact of selective revision in \worksm for requests of different lengths and users with different number of historical posts. We see that revision benefits requests of medium length and users with few posts. From Figure \ref{fig-selective-rev-perq-lens}, we hypothesize that requests that are too short may require additional user input and see no gains from revision. On the other hand requests that are too long, may be more challenging to follow and are unlikely to improve from revisions. From Figure \ref{fig-selective-rev-perq-users}, we see that users with few posts benefit from revision indicating that these users see poorer retrievals. On the other hand users with larger profiles see a drop in performance indicating that even better calibration performance may improve performance of selective revision further.

Note that we don't report results with \texttt{davinci-003} since our procedure for learning a threshold $\theta$ for selective revision failed to find a threshold where dev set performance was improved from selective revision. Finally note that metrics reported for selective revision in Table \ref{tab-both-routing-baseline} isn't directly comparable to those of Tables \ref{tab-both-main-results}, \ref{tab-both-gpt35t-ablations}, and \ref{tab-both-davinci-ablations}  since they represent different LLM runs and exclude a dev set from \worksm and \aita for learning $\theta$ (50 and 200 instances respectively).

\section{Extended Related Work}
\label{sec-related-work-extended}
Having discussed the closest body of related work in \S\ref{sec:related} we discuss additional related work here.

\textbf{Dynamic prompts for LLMs} Besides training retrievers for in-context example selection discussed in \S\ref{sec:related}, other approaches have explored better use of pre-trained models for example selection. \citet{creswell2023selectioninference} select examples based on the target LLM likelihood - necessitating access to LLM likelihoods and incurring latency in retrieval. \citet{gupta2023coveragebased} explore selecting \emph{sets} of examples with dense retrieval models, presenting a complementary approach to ours. Finally, \citet{pan2023knowledgeselector} use retrieval models to select examples from multiple knowledge sources and train a routing model to decide the source of knowledge to retrieve from -- selective revision (\S\ref{sec-routing-results}) based on a retriever may be seen as a form of routing.

\textbf{Prompt robustness in LLMs} Simultaneous routing and retrieval also relates our approach to work ensuring that LLMs are robust to noisy retrievals. Prior approaches ensured robustness by only using retrieved documents based on simple frequency filters on entities mentioned in the input query \citep{mallen2023trustcontext} or based on predictions from an NLI model that determines if the query entails the retrieved contexts \citep{yoran2023making}. Other approaches have sought to fine-tune the LLM to be robust to irrelevant contexts \citep{li2023controlcontext, luo2023sail, yoran2023making} or modify the decoding procedure \citep{shi2023trusting}. In contrast, we determine the quality of the input context based on scale-calibrated retrieval model scores.

\textbf{LLM chaining} In selectively editing generations with an LLM for low-performing requests, our approach also relates to recent work on composing LLMs with other models to build more complex systems \citep{tongshuang2022aichains, arora2023ask, khattab2023demonstratesearchpredict}. Close work is presented by approaches that leverage repeated LLM calls to verify the reasoning or factuality of previous generations \citep{shridhar2023screws, dhuliawala2023chainofverification}. In contrast, our work leverages an efficient retrieval model to selectively direct low-performing generations for further revision, reducing the total number of expensive LLM calls necessary. In this respect, our approach bears similarity to \citet{zhang2023ecoassistant}, who progressively run larger LLMs only when necessary for an input.

\textbf{Calibrated retrievers}
A small body of work has explored calibrated ranking models. \citet{le2022scalecalib} train scale-calibrated ranking models for recommendation models used for advertisement pricing systems. On the other hand, our work leverages scale-calibration for personalized writing assistance. Other work has explored joint training of retrievers and generative models to obtain calibrated retrievers \cite{dhuliawala2022mrcalibration}, using Gaussian embeddings to estimate retriever uncertainty \cite{zamani2023multivar}, or estimating retriever confidence with monte-carlo dropout \cite{cohen2021uncertainrel}. In contrast with probabilistic uncertainty estimation, \proposedmethod minimally modifies training to result in a calibrated model and does not require extensive changes to training, model architecture, or additional inference costs.

\textbf{Writing assistants} A sizable body of work has explored the development of writing assistants. Compared to assistants for communication applications, these have been targeted at authors of creative texts like screenplays \citep{mirowski2023screens}, stories \citep{akoury2020storium}, and poems \citep{goncalo2017poemsurvey} -- consequently, they focus on diverse generations and long-range coherence, rather than personalization. Further, while our work leverages a request-driven assistant, prior systems have used a variety of interaction and control methods. While text completion presents a common form of interaction \citep{clark2018poems}, recent work has seen use of infilling, tag-based control \citep{sun2021iga}, and instruction guided generations \citep{chakrabarty2022instpoem} -- a deeper examination of control and interaction strategies and their trade offs are presented in related reviews \citep{zhang2023ctrlgen, lin2023prompts}. While our approach to personalization may be extended to some alternative interaction paradigms, other interaction techniques are likely to necessitate additional work.

\textbf{Personalized text generation} While we have focussed on author personalization that aims to mimic stylistic patterns, interests, and values of an author, we briefly review reader-personalized text generation -- a setup aiming to generate texts that are engaging and relevant to readers' preferences. This has historically been explored for generating personalized reviews \citep{ni2017estimating}, recipes \citep{majumder2019generating}, news headlines \citep{ao2021pens} and in dialogue agents \citep{mazare2018training, zhang2018personalizeddial}. Related work is also found in text simplification and lay summarization in the context of scientific text -- this work has explored generating definitions for scientific concepts at varying levels of complexity \citep{august2022generating, murthy2022accord} or summarizing scientific text for lay readers \citep{guo2021laysumm}. While recent work has explored this with modern LLMs \citep{li2023large, farajidizaji2023possible}, reader personalization remains an understudied problem and presents a rich area for future work.

\section{Limitations}
\label{sec-limitations}
Here, we discuss limitations of our work derive from our choice of $\slm$ and $\llmG$, our evaluation setup, and the design of our method. 

\textbf{Choice of LLMs} Our experiments use two closed LLMs through API access (\texttt{davinci-003}, \texttt{gpt-35-turbo}).
While we show the value of \proposedmethod with LLM's of varying performance, establishing its effectiveness with other LLMs will require further work. We also acknowledge that closed LLMs limit experimental reproducibility - however, given the widespread use of GPT models \citep{krystal2023chatgptrec} we believe our investigation is meaningful. Finally, in constructing training data for instance selection models for an LLM, prior work has noted the best empirical performance from matching $\slm$ and $\llmG$ \citep{rubin2022promptret}. While we demonstrate benefits from using significantly smaller models for $\slm$, using an open LLM will allow further validation of this result in the context of our approach. However, using a larger (open) model for $\slm$ will incur additional costs in creating training data, and smaller models for $\llmG$ are likely to see a worse generation performance - exploring this tradeoff requires future work. 

\textbf{Evaluation setup} Next, while \worksm represents an impactful and realistic use case for writing assistants, we acknowledge that its private nature limits reproducibility.
Further, our evaluation set of \worksm and \aita represents a limited set of scenarios that are likely to leverage writing assistants. 
While we believe our work represents a meaningful first step, additional future work, and online evaluations are necessary to establish the value of \proposedmethod across the myriad of scenarios where writing assistants may be used. Finally, while we leverage several evaluation strategies to demonstrate the value of \proposedmethod, evaluating text generations under personalization setups represents is an under-explored and a currently emerging body of work \citep{wang2023personalizedeval, wang2023automated}.

\textbf{Method design} Finally, we note that the current design of \proposedmethod is likely to have some drawbacks. It is possible that our proposed method for training instance selection biases system performance toward some users or requests -- we leave examination of this to future work. It is also possible that formulating $\retriever$ as an expressive crossencoder and the use of large LLMs will present latency limitations for interactive applications -- exploration of models supporting faster retrieval and text generation inference represent important future work.

\begin{lstfloat*}[t]
\begin{lstlisting}[breakautoindent=false, breaklines=true, breakindent=0pt]
## TASK
Given an enterprise social media post, generate a set of writing instructions that explain how to
"reverse-engineer"; the post. Use the following steps:
- The instructions must give a high-level overview of what the post aims to communicate. Example: [readcted]
- The instructions must include specific proper nouns (people, places, organzations). Example: [redacted]
- The instructions must include contact information if available. Example: [redacted]
- The instructions must include specific links to websites or files if available. Example: [redacted]
- The instructions must contain any knowledge that is highly specialized and is likely to be only known to the individual who wrote the post, if available. Example: [redacted]
- The instructions must contain rough sketches of any personal anecdotes in the post, if available. Example: [redacted]
- The instruction must **not** contain any formatting or ordering information from the post.

## OUTPUT
Output the following:
<Explanation>{explanation of your reasoning for how you generated the instructions, in 3 sentences or fewer}</Explanation>
<Instruction.Overview>{1-2 sentences overview of what the post aims to communicate}</Instruction.Overview>
<Instruction.Names>{1-2 sentences about the people, places, or organizations mentioned in the post, _NONE_ if not applicable}</Instruction.Names>
<Instruction.Contacts>{1-2 sentences about the contact information copied verbatim in the post, _NONE_ if not applicable}</Instruction.Contacts>
<Instruction.Links>{1-2 sentences including the links copied verbatim from the post, _NONE_ if not applicable}</Instruction.Links>
<Instruction.Knowledge>{1-2 sentences paraphrasing the specialized knowledge included in the post, _NONE_ if not applicable}</Instruction.Knowledge>
<Instruction.Anecdotes>{1-2 sentences paraphrasing the anecdotes included in the post, _NONE_ if not applicable}</Instruction.Anecdotes>

## INPUT
+\textcolor{Fuchsia}{\textbf{\{\{input\_post\}\}}}+
\end{lstlisting}
\caption{GPT4 prompt used to generate synthetic requests for \worksm posts in our training set.}
\label{fig-worksm-reqgen}
\end{lstfloat*}
\begin{lstdoc*}[t]
\begin{lstlisting}[breakautoindent=false, breaklines=true, breakindent=0pt, backgroundcolor=\color{gray!10}, basicstyle=\ttfamily]
+\textbf{Overview}+: 
In this study, we are developing LLM-based approaches for writing social media posts on enterprise social networks. Your task is as follows: Given a social media post from an enterprise social media platform, write a short outline of the post. In writing your outline, imagine you are a manager, social media manager, or event organizer writing a rough sketch of the post with the key information you would like to share.

+\textbf{Data Format}+: 
You are given a spreadsheet consisting of ~150 English posts. Each row corresponds to a single post. The spreadsheet contains the following columns: PostId, InputPost, OutputShortOutline. The first column is the ID of the post; you can ignore this column. The second column is the full text of the input post. In the third column, you will write your short outline based on the input post.

+\textbf{DO's for your outline}+:
When writing your short outline, +\ul{do}+ include the following:
- One sentence about the goal of the post: Include a brief description of what the post is trying to communicate. Example: [redacted]
- Specific proper nouns (people, places, things): Include names of specific people, places, or things in your outline. Example: [redacted]
- Specialized knowledge: If the knowledge contained in the post is highly specialized and is likely to be only known to the individual writing the post, include a rough sketch of that information in your outline. Example: [redacted]
- Personal anecdotes: If the post contains specific personal anecdotes, include a rough sketch of that information in your outline. Example: [redacted]
- Special emphasis or call to action: If the post makes a special emphasis, include a rough sketch of that emphasis or call to action in your outline. Example: [redacted]
- External website links: If the post links to an external website, include the link in your outline. Example: [redacted]

+\textbf{DONT's for your outline}+:
When writing your short outline, +\ul{do not}+ include the following:
- Anything related to the ordering of content.
- Formatting instructions.
- Any verbatim text other than specific proper nouns.
\end{lstlisting}
\caption{Instructions provided to annotators for authoring requests for our evaluation set in \worksm.}
\label{fig-req-guidelines}
\end{lstdoc*}

\end{document}